\documentclass[final,3p,times]{elsarticle}

\usepackage{amssymb}
\usepackage{soul}
\usepackage{color}
\usepackage{url}
\usepackage{hyperref}
\usepackage{float}
\usepackage{subcaption}
\usepackage{amsmath}
\usepackage{multirow}
\usepackage{fancyhdr}

\usepackage{lineno}

\journal{Neurocomputing}

\begin{document}

\begin{frontmatter}



\title{Enhancing Object Detection for Autonomous Driving by \\ Optimizing Anchor Generation and Addressing Class Imbalance
}


\author{Manuel Carranza-García\corref{cor1}}
\cortext[cor1]{Corresponding author.}
\ead{mcarranzag@us.es}
\author{Pedro Lara-Benítez}
\author{Jorge García-Gutiérrez}
\author{José C. Riquelme}

\address{Division of Computer Science, University of Sevilla, ES-41012 Seville, Spain}


\begin{abstract}
Object detection has been one of the most active topics in computer vision for the past years. Recent works have mainly focused on pushing the state-of-the-art in the general-purpose COCO benchmark. However, the use of such detection frameworks in specific applications such as autonomous driving is yet an area to be addressed. {This study presents an enhanced 2D object detector based on Faster R-CNN that is better suited for the context of autonomous vehicles. Two main aspects are improved: the anchor generation procedure and the performance drop in minority classes.
The default uniform anchor configuration is not suitable in this scenario due to the perspective projection of the vehicle cameras. Therefore, we propose a perspective-aware methodology that divides the image into key regions via clustering and uses evolutionary algorithms to optimize the base anchors for each of them. Furthermore, we add a module that enhances the precision of the second-stage header network by including the spatial information of the candidate regions proposed in the first stage. We also explore different re-weighting strategies to address the foreground-foreground class imbalance, showing that the use of a reduced version of focal loss can significantly improve the detection of difficult and underrepresented objects in two-stage detectors.}
Finally, we design an ensemble model to combine the strengths of the different learning strategies. Our proposal is evaluated with the Waymo Open Dataset, which is the most extensive and diverse up to date. The results demonstrate an average accuracy improvement of 6.13\% mAP when using the best single model, and of 9.69\% mAP with the ensemble. The proposed modifications over the Faster R-CNN do not increase computational cost and can easily be extended to optimize other anchor-based detection frameworks.
\end{abstract}



\begin{keyword}
autonomous vehicles \sep anchor optimization \sep class imbalance \sep convolutional neural networks \sep deep learning \sep object detection

\end{keyword}

\end{frontmatter}

\renewcommand{\headrulewidth}{0pt}
\thispagestyle{fancy}
\lhead{\footnotesize{\noindent When referring to this paper, please cite the published version: M. Carranza-García, P. Lara-Benítez, J. García-Gutiérrez, J.C. Riquelme,
Enhancing Object Detection for Autonomous Driving by Optimizing Anchor Generation and Addressing Class Imbalance. \textit{Neurocomputing} (2021).  \url{https://doi.org/10.1016/j.neucom.2021.04.001}}}.



\section{Introduction}
\label{intro}

Developing robust machine learning models that can accurately detect and classify multiple objects in an image remains a core challenge in computer vision. Object detection has attracted the interest of many researchers due to its application to multiple real-world problems such as autonomous driving \cite{Alvaro:2018, Yin:2020b}, robotic vision \cite{Hoseini:2019}, security surveillance \cite{Salazar:2020}, or land monitoring \cite{Carranza:19}. In recent years, the latest advancements in this field have been achieved thanks to the development of deep convolutional networks. Deep learning has proven to be a very powerful tool for learning abstract hierarchical representations of raw input data \cite{leCun:2015}. With the increase of availability and quality of remote sensing data collected by different sensors (higher resolution RGB cameras, LiDAR and radar data, etc.), deep learning models have pushed the state-of-the-art in many visual recognition tasks \cite{Hassaballah:2020}. 

In particular, object detection is one of the main perception problems that the advanced driver assistance system (ADAS) of autonomous vehicles faces. The multi-modal sensors equipped in these vehicles provide valuable data to be used with popular deep learning-based object detectors. Nevertheless, the perception systems of self-driving vehicles need to be accurate and robust enough to operate safely in complex scenarios such as mixed urban traffic, adverse weather conditions, unmapped roads, or areas with unreliable connectivity \cite{Litman:2020}. Under these circumstances,  it is still hard for existing detectors to perceive all traffic participants (vehicles, pedestrians, traffic signs, etc.) accurately, robustly, and in real-time. The impact of autonomous driving in the future promises to be important due to their potential to improve road safety, reduce traffic, and decrease pollution \cite{Liu:2019}. However, many aspects require significant progress before this technology can fully substitute human driving.

In this work, the aim is to enhance the 2D object detection accuracy in the images obtained from the on-board cameras of autonomous vehicles. The goal of this detection task is to determine the presence of objects from given categories and return the spatial location of each instance through a bounding box \cite{Liu:2019b}. In recent literature, the main trend in object detection is to develop increasingly sophisticated architectures to improve the performance over the general-purpose COCO (Common Objects in Context) benchmark \cite{Coco:2014}. However, the effectiveness of such generic object detectors when applied to particular applications is still far from optimal \cite{Zhao:2019}. For this reason, {this study proposes several modifications} to the popular Faster R-CNN detection framework to better adapt it to the specific context of self-driving cars. {The novelty of this work lies in two main aspects that are considered for improving the performance of the original model: optimizing the anchor generation procedure and modifying the learning process to improve accuracy over minority instances.}

The uniform anchor generation procedure of Faster R-CNN is not suitable for the autonomous driving scenario. The default configuration, which has proven to be effective for generic object detection, produces anchors with the same scale and aspect ratios at each location of the feature map. However, due to the perspective projection of on-board cameras in the vehicles, the scale of objects has a strong correlation with their position in the image in this context. This implies that in regions where objects tend to be very large, producing small scale anchors is not appropriate, and vice versa. To overcome this anchor mismatch issue, our proposal is to divide the images into several regions and optimize each of them independently. With the help of a clustering study, key regions in the images that have objects with significantly different dimensions {are obtained}. Then, a methodology based on evolutionary algorithms {is presented} in order to search for optimal values of scale and aspect ratio for the prior anchors of each region. Furthermore, we modify the second-stage header network introducing spatial properties extracted from the region of interest (ROI) proposals of the first-stage. The spatial features of ROIs (size and position in the image) {are concatenated} to the convolutional features extracted from the backbone network to improve localization accuracy.

Another important issue is that the default training scheme of Faster R-CNN results in a significant performance drop in minority classes and difficult instances. In the literature, the learning process of detectors has been given less attention compared to the development of architectures \cite{Pang:2019:Libra}. However, a balanced learning scheme is crucial for this multi-class scenario due to the presence of elements, such as pedestrians or cyclists, that are less frequent than vehicles. In this study, an extensive analysis of different approaches to address {foreground-foreground class imbalance is performed}. Several alternatives {are explored}, such as assigning different weights according to the class distribution and the use of focal loss, which has been traditionally used in one-stage detectors. Finally, an ensemble model based on non-maximum suppression and test-time augmentation {is designed}, combining the different training strategies to increase the robustness of the detector.

The recently released Waymo Open Dataset \cite{Waymo:2019} {is used} for evaluating the proposal, which is the largest and most diverse up to date in terms of geographic coverage and weather conditions. Waymo is 15 times more diverse than other existing benchmarks such as KITTI \cite{Kitti}. In this dataset, the object detection task has objects from three classes (vehicles, pedestrians, and cyclists) that are divided into two difficulty levels. To the best of our knowledge, this is the first study that addresses anchor optimization and class imbalance in a multi-class 2D detection problem in the context of autonomous vehicles. The proposed methodology can easily be extended to other anchor-based detection frameworks with different backbone networks, since it does not rely on the specific implementation carried out in this study.

In summary, the main contributions of this work can be compiled as follows:
\begin{itemize}
    \item {A novel region-based anchor optimization methodology using evolutionary algorithms for 2D object detection for autonomous driving}
    \item {A module that enhances the second-stage header network of Faster R-CNN by including the spatial features of ROIs produced by the region proposal network.}
    \item {A thorough study of different training procedures to address  severe foreground-foreground class imbalance in two-stage object detectors.}
    \item {An ensemble model combining the strengths of the different learning strategies to improve detection precision.}
\end{itemize}

The rest of the paper is organized as follows: Section 2 presents a review of related work; in Section 3 the materials used and the methods proposed in the study are described; Section 4 reports and discusses the results obtained; Section 5 presents the conclusions and potential future work. 


\section{Related work}

Recent progress in the object detection field has been driven by novel methodologies based on deep learning, as it has happened in many computer vision tasks \cite{Liu:2019:CBNet}. Existing image object detectors in the literature can be mainly divided into two categories: two-stage detectors such as Faster R-CNN \cite{Ren:2017}, and one-stage detectors such as SSD \cite{Liu:2016:SSD}. Generally speaking, the strength of one-stage detectors lies in their higher inference speed, while two-stage architectures obtain higher localization accuracy. 

The pioneering two-stage detector was the Regions with CNN features framework (R-CNN) \cite{Girshick:2014}. R-CNN used the selective search method to crop box proposals from the image and feed them to a convolutional network classifier. This external proposal generation was very costly and inefficient. Faster R-CNN solved this issue by sharing features between the region proposal network (RPN) and the detection network \cite{Ren:2017}. 
This approach improved accuracy and speed and has led to a large number of follow-up works. For instance,  R-FCN proposed a position-sensitive ROI cropping that respects translation variance \cite{Jifeng:2016:RFCN}. Later studies have focused on exploiting the multi-scale properties of feature extractors, such as the feature pyramid networks (FPNs) with lateral connections proposed in \cite{Lin:2017:FPN}. 
{Other works have tried to improve this detector by including rotation-invariant
and Fisher discriminative regularizers on the CNN features.} \cite{Cheng:2019} 
{Due to the cost of manually labeling bounding boxes, there are also many studies on weakly-supervised object detection that work only with image-level labels} \cite{Cheng:2020}. 
{Currently, the COCO leaderboard is led by Cascade R-CNN methods. Cascade models build a sequence of detectors trained with increasing intersection-over-union (IoU) thresholds, to be sequentially more selective against false positives} \cite{Cai:2018:Cascade}.

In contrast, one-stage architectures predict class probabilities and bounding box offsets directly from the image, without the region proposal step. The first YOLO (You Only Look Once) architecture was proposed in \cite{Redmon:2016}, achieving real-time inference rates but with high localization errors. 
SSD combined ideas from YOLO and RPN to improve the performance while maintaining high speed \cite{Liu:2016:SSD}. With the help of default bounding boxes, SSD detects objects at different scales on several feature maps. More recently, an important step forward in the one-stage family was achieved by solving the foreground-background class imbalance problem. RetinaNet proposed a novel focal loss function that focuses on difficult objects by down-weighting the importance of well-classified samples  \cite{Lin:2020:Focal}. 
{Another interesting approach are the anchor-free one-stage detectors, that do not require pre-defined anchor boxes. FCOS} \cite{FCOS:2019} {and CenterNet} \cite{CenterNet:2019} {use the center of objects to define positives and regress the four distances that build the bounding box from that point. Other models such as ExtremeNet} \cite{Zhou2019:ExtremeNet} {or CornerNet} \cite{Law:2020:CornerNet} {generate the boxes by locating several keypoints first. Anchor-free detectors are more flexible than RetinaNet and can achieve similar performance.}

The backbone network that acts as a feature extractor and its capacity to extract quality features play a very important role in all detection frameworks. Rather than the VGG network \cite{Simonyan:2014:VGG} used in the original Faster R-CNN paper, deeper and more densely connected architectures have been recently proposed. Some examples include the ResNet \cite{He:2016:ResNet} used in the Mask R-CNN detector \cite{He:2017:Mask}, ResNeXt \cite{Xie:2017:ResNeXt}, Res2Net \cite{Gao:2019:Res2Net} or HRNet \cite{Wang:2020:HRNet}. {The improved YOLOv3 detector includes multi-scale predictions and a new feature extractor, DarkNet-53, which uses residual blocks and skip connections} \cite{Yolov3:2018}. Other works such as NAS-FPN \cite{Ghiasi:2019:NAS} introduce neural architecture search to learn optimal feature fusion in the pyramid and build a stronger backbone. However, these complex networks lead to slower inference speed. Since this is undesired in real-time applications, other researchers have focused on designing lightweight backbones such as MobileNets \cite{Sandler:2018:MobileNet}, which are also less accurate. In general, finding the optimal speed/accuracy balance in a backbone architecture is a difficult task that highly depends on the problem to be addressed \cite{Licheng:2019}.

The interest in the autonomous driving field has risen significantly in recent years. Many high-quality datasets are becoming available for the research community to push the state-of-the-art in problems such as object detection. After the popular KITTI benchmark \cite{Kitti}, other datasets have been released such as NuScenes \cite{NuScenes:2019} or PandaSet \cite{PandaSet:2019}. A complete overview of existing self-driving datasets is provided in \cite{Feng:2019}. Waymo has recently released the most extensive and diverse multi-modal dataset up to date \cite{Waymo:2019}.  
{In} \cite{Carranza:2021}, {the authors evaluate the trade-off between accuracy and speed of several state-of-the-art detectors over the Waymo dataset.} However, there are still few works that have addressed the optimization of existing 2D object detectors in the context of autonomous vehicles. RefineNet proposed extra regressors to further refine the candidate bounding boxes for vehicle detection \cite{Rajaram:2016}. The work in \cite{Wang:2019} presents an anchor optimization methodology and ROI assignment improvement over two-stage detectors but also focusing only on vehicles and not on other traffic participants. {A CNN-based methodology to improve vehicle detection in adverse weather conditions was presented in} \cite{Hassaballah:2020b}. {There are other important perception problems in autonomous driving being addressed with deep learning such as 3D detection using LiDAR point clouds} \cite{Yin:2020, Meng:2020} {and object tracking} \cite{Liang:2020,Dong:2019}.










\section{Materials and Methods}
\label{section3}
This section presents the dataset used for the study and the methodology proposed to enhance object detection in the context of autonomous vehicles. Firstly, the anchor optimization procedure and the modifications applied to the Faster R-CNN architecture {are described}. Secondly, the different learning strategies studied to address class imbalance {are explained}. Thirdly, the proposed ensemble model {is presented}. In the final section, the remaining implementation details {are provided} to allow reproducibility. The complete source code can be found at \cite{github-code}.

\subsection{Waymo Open Dataset}
\label{dataset}
\begin{figure}[H]
    \centering
    \includegraphics[width=\textwidth]{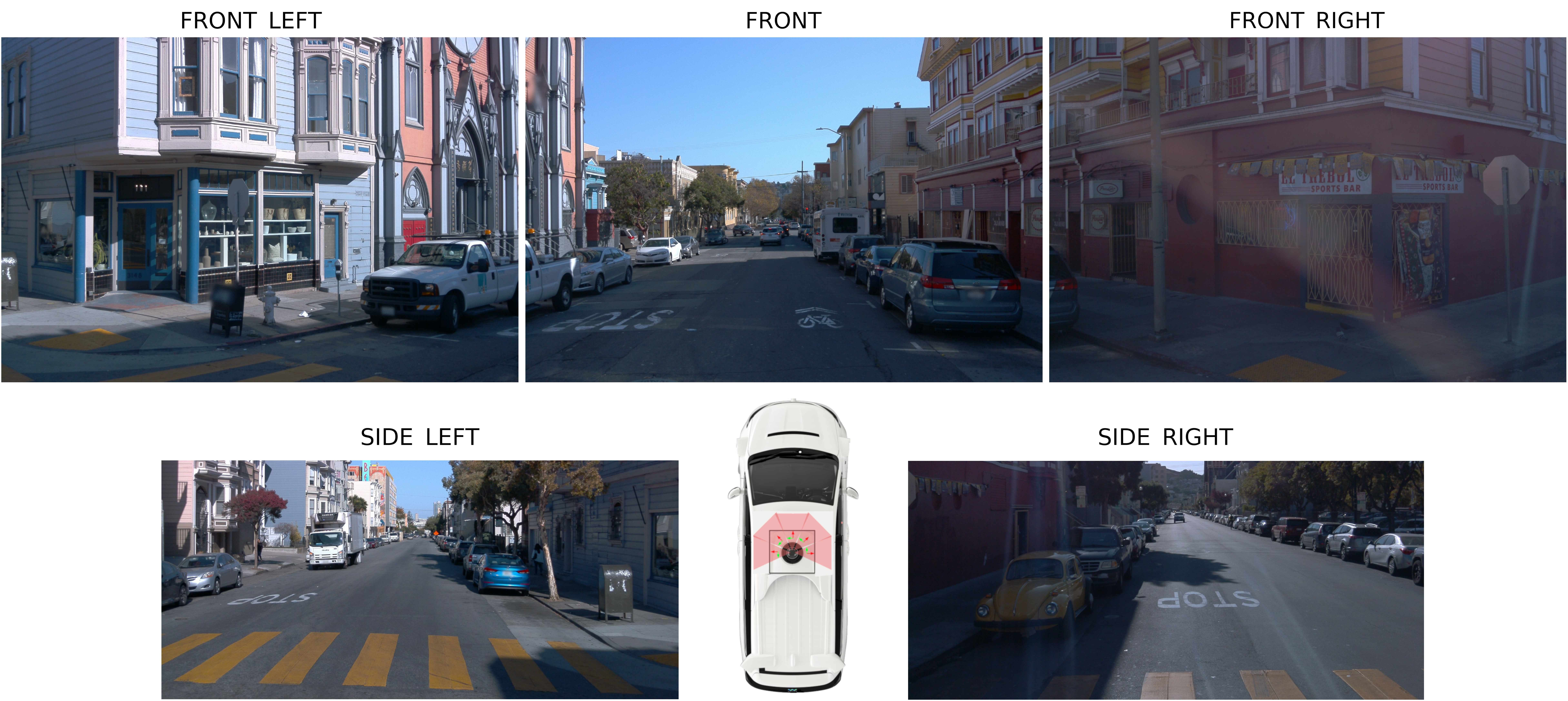}
    \caption{Images from a single frame obtained by the five cameras of Waymo's vehicle}
    \label{fig:five-cameras}
\end{figure}

The Waymo Open Dataset \cite{Waymo:2019} consists of 1150 driving video scenes across different urban areas (Phoenix, San Francisco, and Mountain View) and at different times of the day (day, night, and dawn). Each scene captures synchronized LiDAR and camera data for 20 seconds, resulting in around 200 frames per scene. The problem addressed in this study is 2D object detection. This task is to assign 2D bounding boxes to objects that are present in a single RGB camera image. 
Waymo's vehicle is equipped with five high-resolution cameras (Front, Front Left, Front Right, Side Left, and Side Right). Frontal cameras obtain images with a resolution of $1920x1280$, while lateral cameras have an image size of $1920x886$. All cameras have a $\pm 25.2^o$ horizontal field of view (HFOV). An example of the images captured by all cameras in a single frame is presented in Figure \ref{fig:five-cameras}. As it is done in Waymo's online challenge, the images obtained from all cameras are considered a single dataset for evaluation purposes.


\clearpage

\begin{figure}[H]
    \begin{subfigure}{.498\textwidth}
    \centering
    \includegraphics[width=\textwidth]{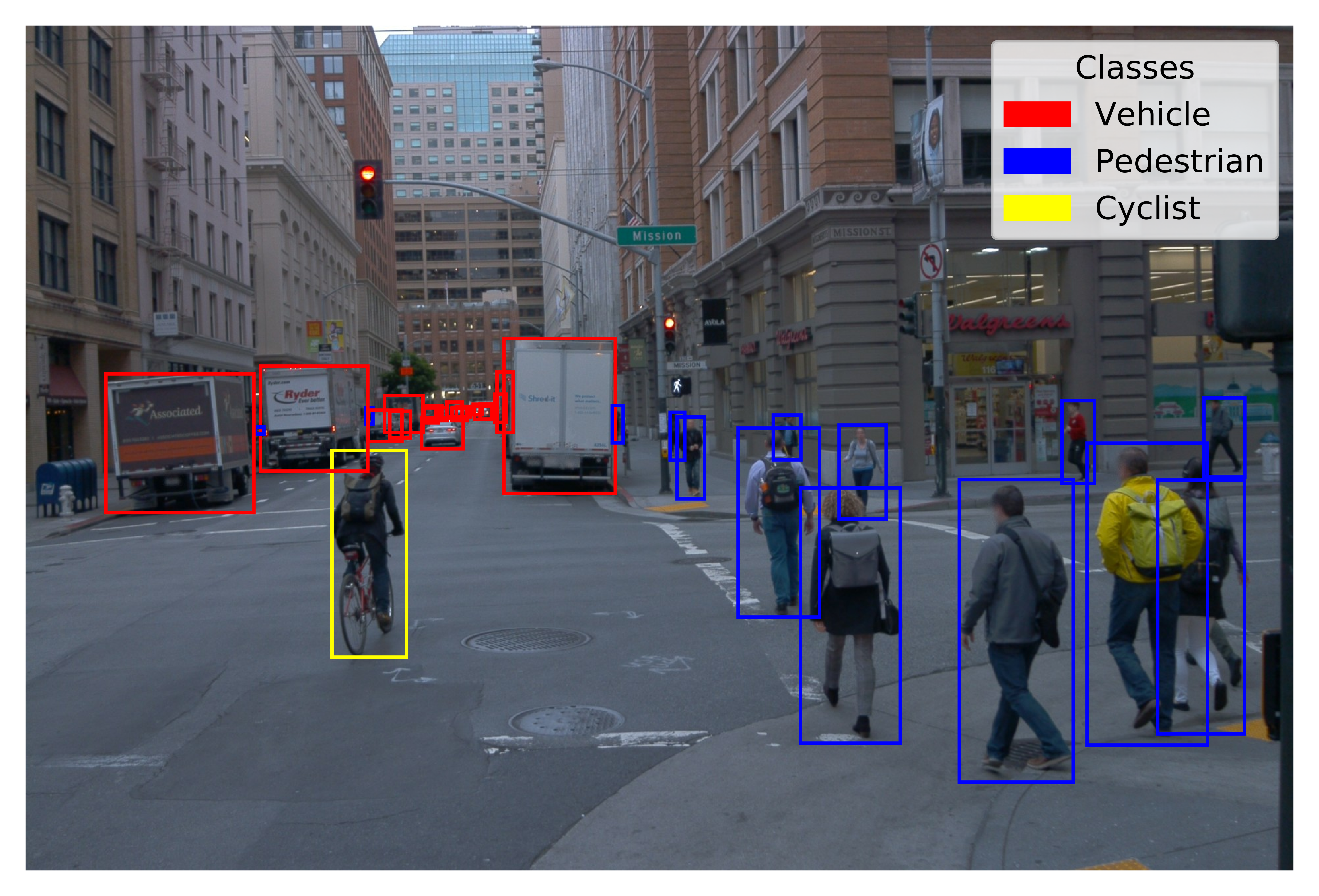}      \caption{Objects of the three different classes}
      \label{fig:waymo-boxes}
    \end{subfigure}
    \begin{subfigure}{.498\textwidth}
      \centering
    \includegraphics[width=\textwidth]{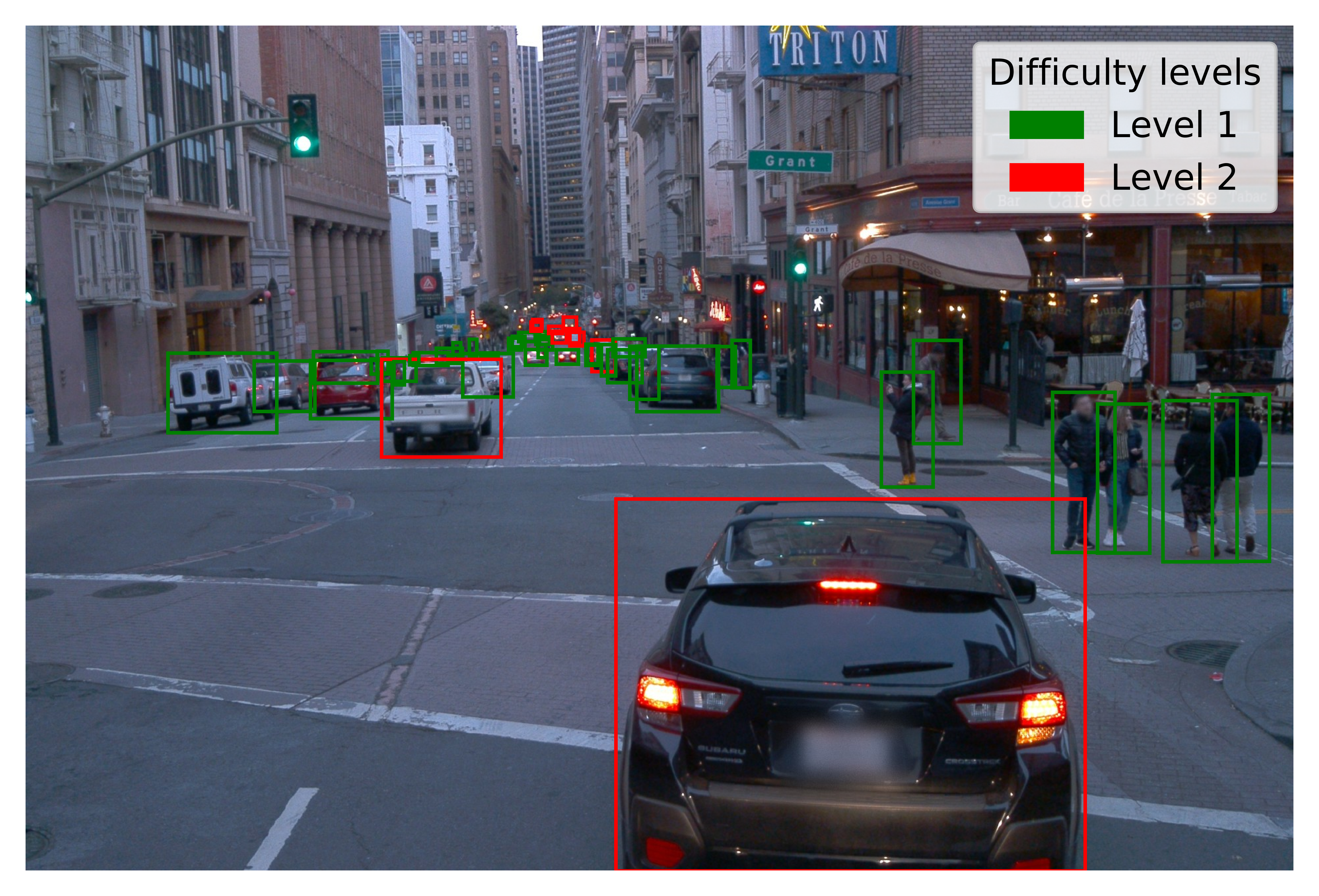}
      \caption{Objects of both difficulty levels}
      \label{fig:waymo-difficulty}
    \end{subfigure}
    \caption{2D object detection problem in the Waymo Open Dataset}
    \label{fig:my_label}
\end{figure}

The dataset contains around 10 million manually annotated labels across all cameras. Three different classes are considered for this problem: vehicles (which includes any wheeled motor object such as cars or motorbikes), pedestrians and cyclists. Figure \ref{fig:waymo-boxes} shows an example of the labeled data provided, which are tightly fitting bounding boxes around the objects. {Furthermore, Waymo provides two different difficulty levels for the labels (Level 1 and 2), which are illustrated in Figure} \ref{fig:waymo-difficulty}. {Level 2 instances are objects considered as hard and the criteria depends on both the human labelers and the object statistics.} For the evaluation, the level 2 metrics are cumulative and also include all objects belonging to level 1. The count of objects of the different classes is presented in Table \ref{tab:label-count}. As can be seen, there is a significant class imbalance between vehicles and pedestrians, and the number of cyclist labels is minimal.

\begin{table}[H]
\centering
\caption{Count of labeled objects in the Waymo dataset}
\label{tab:label-count}
\begin{tabular}{cccc}
\hline
                    & \textbf{Vehicle} & \textbf{Pedestrian} & \textbf{Cyclist} \\ \hline
\textbf{Count}      & 7.7M             & 2.1M                & 63K              \\
\textbf{Percentage} & 78.07\%          & 21.29\%             & 0.64\%           \\ \hline
\end{tabular}
\end{table}

The dataset is divided into 1000 scenes for training and validation (around 1 million images), and 150 for testing (around 150k images). Waymo provides an online submission tool to evaluate the models over the testing set, since those labels are not publicly available. The scenes composing the test set are from a different geographical area, which ensures that the capacity of generalization of trained models is properly evaluated.

\subsection{Faster R-CNN architecture}

\begin{figure}[h]
    \centering
    \includegraphics[width=\textwidth]{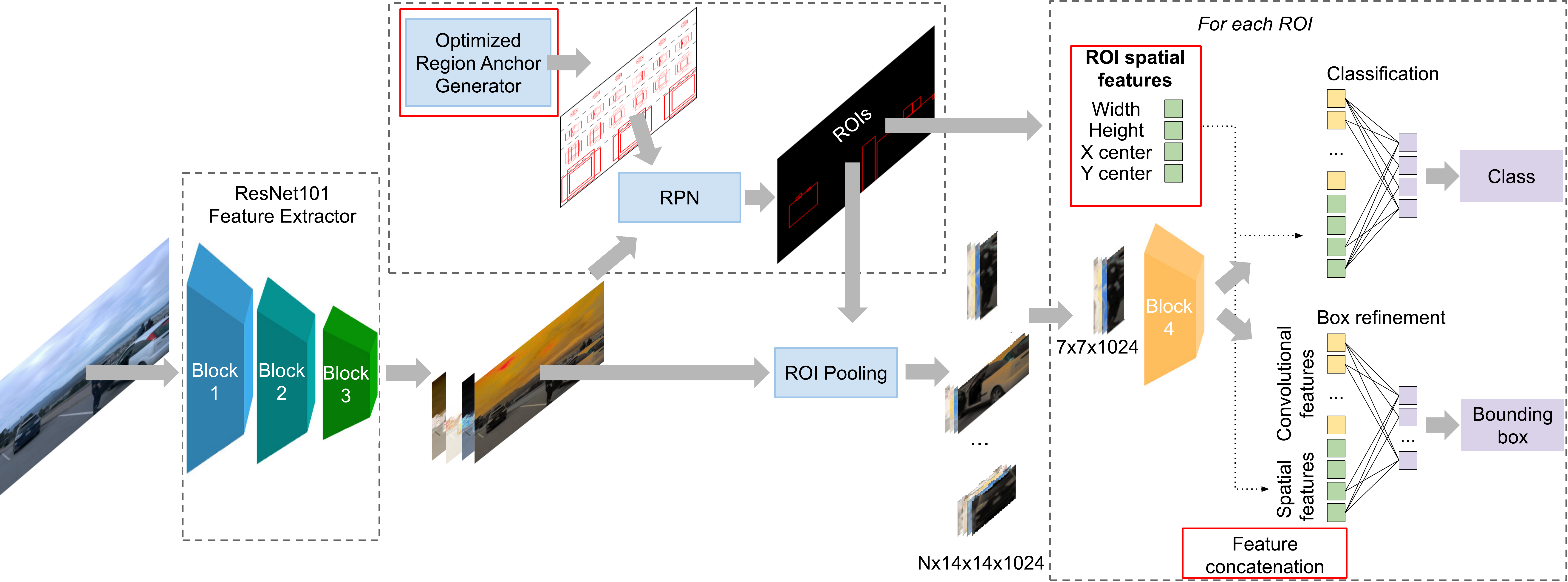}
    \caption{Faster R-CNN architecture. The improvements proposed in this study are highlighted in red. $N$ refers to the number of ROIs passing to the second stage.}
    \label{fig:fasterrcnn}
\end{figure}

Faster R-CNN has been extensively used in the recent literature as a general-purpose object detection framework \cite{Ren:2017}. This detector follows a multi-task learning procedure, combining classification and bounding box regression to solve the detection problem. It uses a convolutional backbone  (e.g. VGG, ResNet) to extract hierarchical features from the images, and consists of two stages: a region proposal network (RPN) and a Fast R-CNN header network. Figure \ref{fig:fasterrcnn} shows the Faster R-CNN architecture, illustrating the complete two-stage process.

In the first stage, the RPN uses features from an intermediate level of the feature extractor to predict class-agnostic box proposals (object or background). This is achieved by predicting multiple candidate boxes at each location using multi-scale reference anchors. Afterwards, a limited number of these proposals (typically 300) are selected as regions of interest (ROIs) and pass to the second stage. The selected ROIs are used to crop features from the same intermediate feature map using a ROI pooling operation. Those cropped features are then fed to the remaining layers of the backbone network to predict a class and perform a box refinement for each proposal. In Faster R-CNN, convolutional features are shared between both stages, which improves accuracy and speed. However, its bottleneck is the number of ROIs proposed by the RPN, since the computation of the second stage is run once per each proposal.

In this study, the ResNet-101 backbone network is used, since it provides a good speed/accuracy trade-off \cite{Huang:2017:Tradeoffs}. It is also the network provided for the baseline results published with the Waymo dataset, which allows a fair comparison. As in the original implementation \cite{He:2016:ResNet}, features used in the RPN are extracted from block 3 of the ResNet. In the second stage, ROIs are cropped and resized to $14x14$, and then max-pooled to $7x7$ before being fed to block 4. However, our proposal does not rely on this specific backbone and can be used with other existing anchor-based two-stage detectors.

Despite its advantages, there are several aspects of Faster R-CNN that can be improved to better adapt it to the characteristics of the problem addressed in this study, which is object detection in the autonomous driving scenario. The improvements proposed to the original architecture are highlighted in red in Figure \ref{fig:fasterrcnn}:  the per-region anchor generation optimization in the RPN; and the ROIs spatial features concatenation in the header classification network. The details of these modifications are provided in the following sections.

\subsection{Anchor generation optimization}
\label{anchor-opt}
The first stage of the Faster R-CNN is the region proposal network (RPN) that selects the ROIs to be forwarded to the detection stage. The ROIs are generated using a convolutional sliding window over some intermediate feature map of the backbone network. These proposals are parametrized relative to reference boxes known as anchors. In order to detect different sized objects, the RPN predicts multiple region proposals at each location by using multi-scale anchors. There are $k$ anchor boxes with different scales and aspect ratios centered at each pixel of the feature map. The scale ratio is defined with respect to a base of 256, and the aspect ratio is the width over the height of the box. The size and shape of these anchors have to be manually defined and is critical for the success of the detector \cite{Ahmad:2020}. 

In this self-driving scenario, the shape of objects captured by the cameras can be significantly different depending on their class and position in the image. The shape of pedestrians tends to be tall and narrow, while vehicles are often wider and more squared. Furthermore, the perspective projection of the cameras equipped in autonomous vehicles is an important factor that must be considered for object detection \cite{Wang:2019}. The original RPN implementation in Faster R-CNN using ResNet proposes multiple anchors with different scales ratios $(0.25, 0.5, 1, 2)$ and aspect ratios $(0.5, 1, 2)$ at each location \cite{He:2016:ResNet}. This uniform configuration was found to work well for general-purpose object detection, but it is far from optimal in this particular application. Those pre-defined scales and aspect ratios do not coincide with the size of the objects seen from on-board cameras, hence resulting in many invalid anchors and useless computation. Therefore, in this study, the aim is to find a better configuration for those values of scale and aspect ratio.

\begin{figure}[H]
    \centering
    \includegraphics{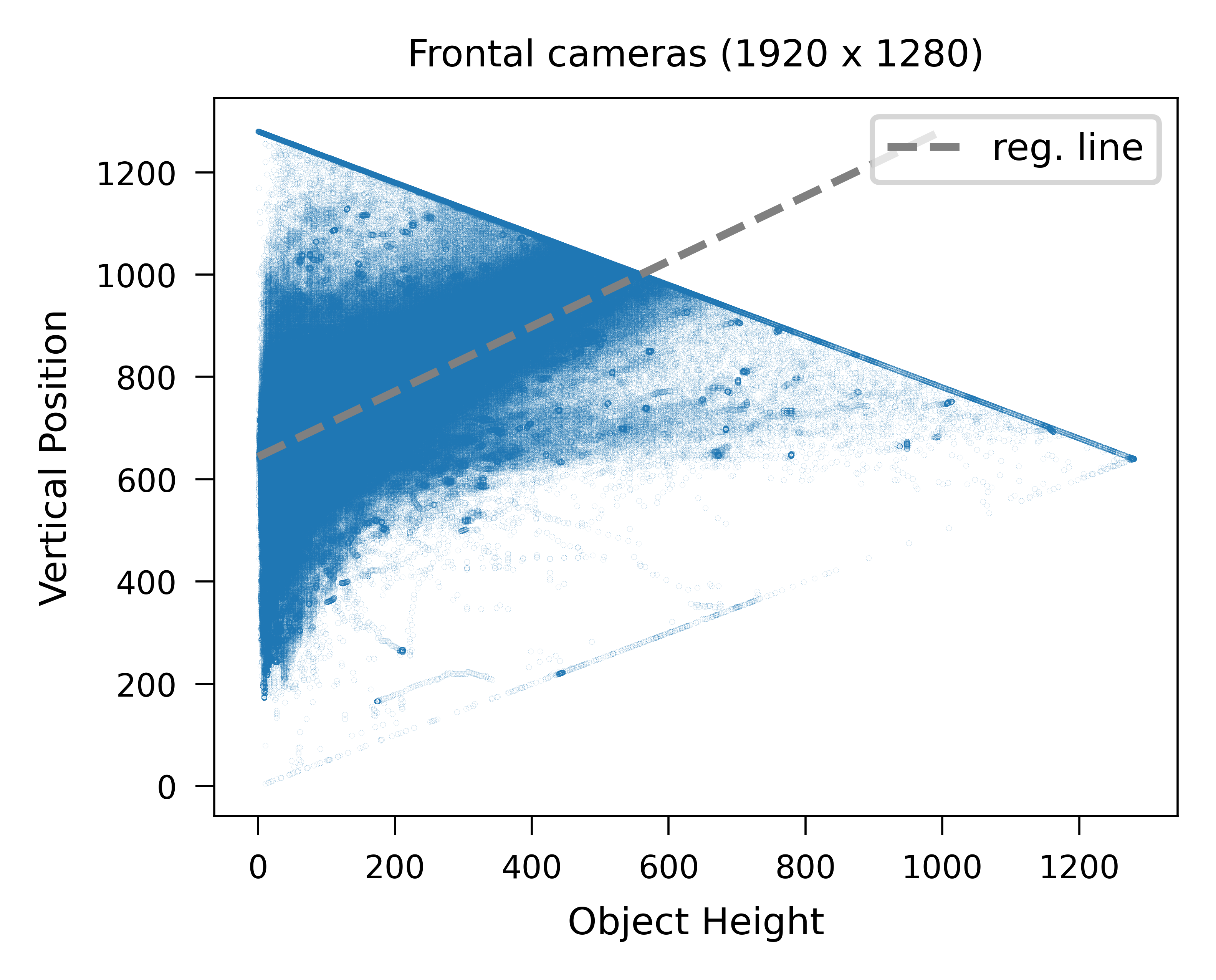}
    \includegraphics{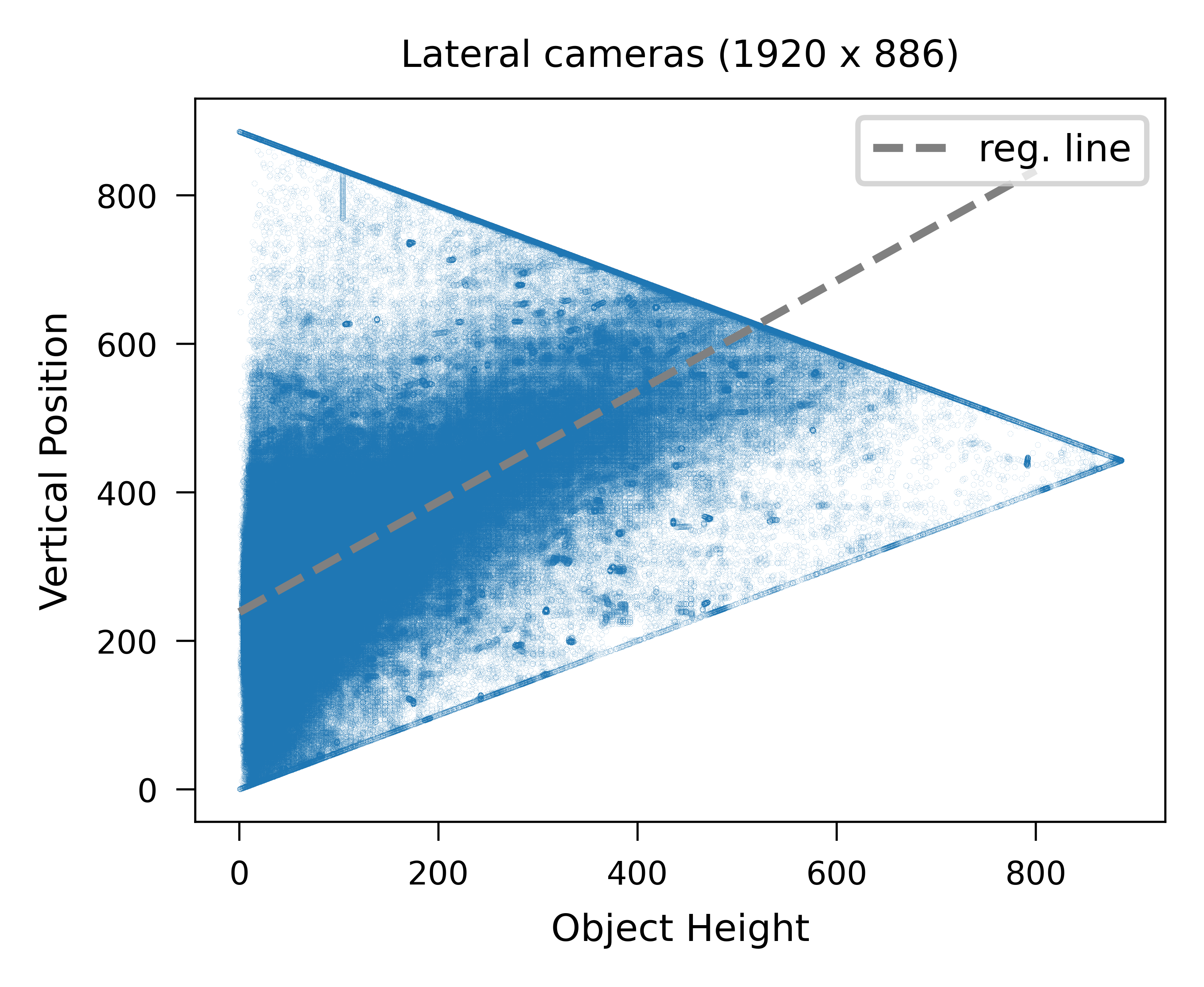}
    \caption{Correlation between the object size and its location in the image. The (0,0) point refers to the top left corner of the image. The regression line illustrates the strong positive correlation.}
    \label{fig:correlation}
\end{figure}

Due to the perspective of the cameras in the vehicle, the size of the captured objects highly depends on their position in the image. Therefore, one of the first steps of this study is to analyze the relationship between bounding box dimensions and their location in the image. Figure \ref{fig:correlation} displays the distribution of all objects in the dataset with respect to two variables: the vertical position of the center of the object (y-axis center), and the object's height. Due to the different dimensions and nature of frontal and lateral cameras, the analysis is separated for both of them to check for any significant particularities. As can be seen in the figure, there is a strong correlation between both variables. For frontal cameras, there is a 0.67 Pearson correlation coefficient, while for lateral cameras the correlation is 0.69. This positive correlation implies that objects of larger size tend to appear at the bottom part of the image, while smaller objects are more often at the top part of the image, as they are further away.

These findings confirm the fact that a uniform anchor generation across the image is not optimal for this context.  This anchor dimension mismatch can significantly decrease the performance of the detector. The next steps of the paper focus on how the anchor generation process has been modified to adapt it for this specific problem and improve the detection accuracy. Our perspective-aware proposal is divided into two steps: the division of the images in key regions using a clustering analysis, and the per-region anchor optimization using an evolutionary algorithm.


\subsubsection{Region division using clustering}

\begin{figure}[H]
\centering
    \includegraphics{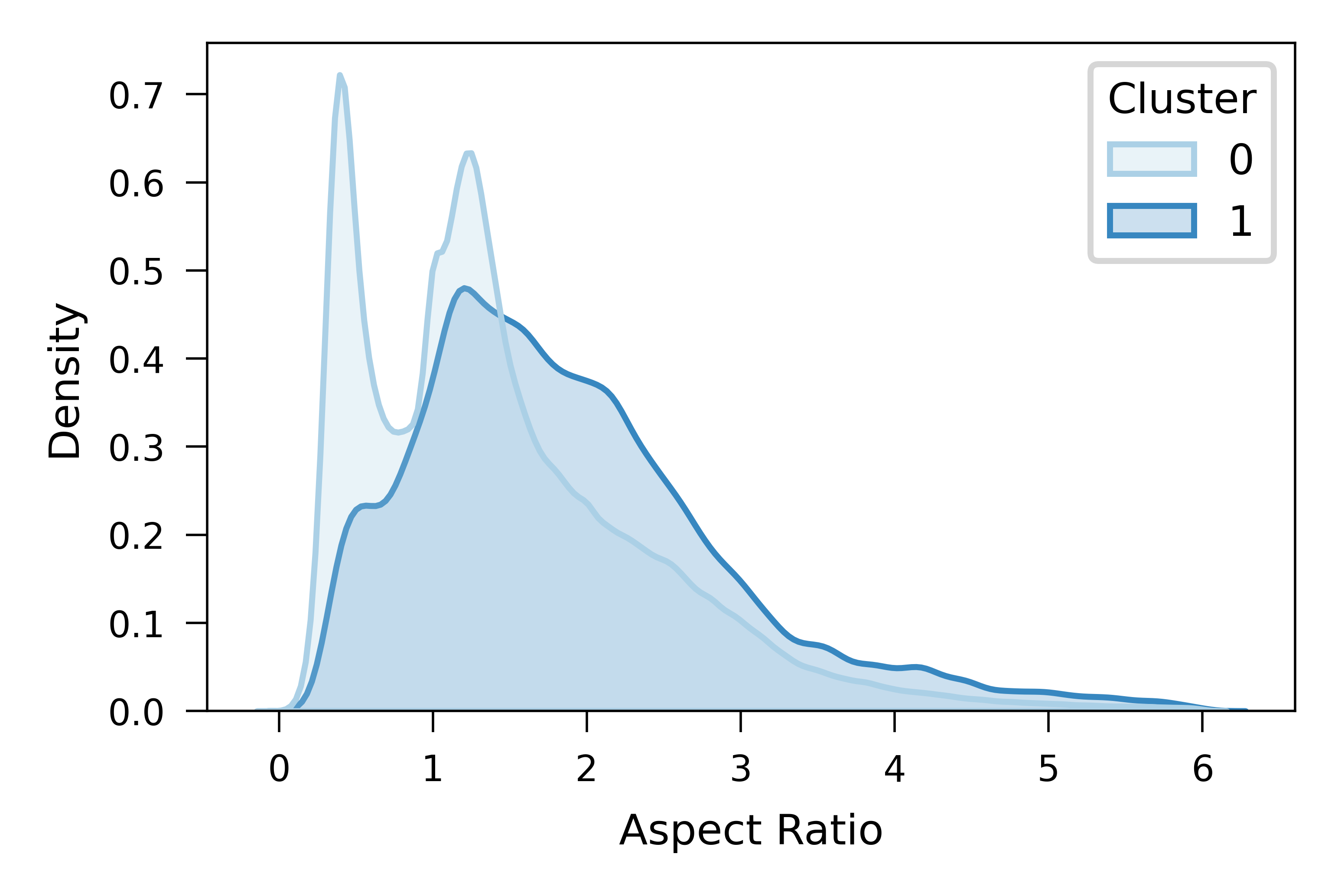}
    \includegraphics{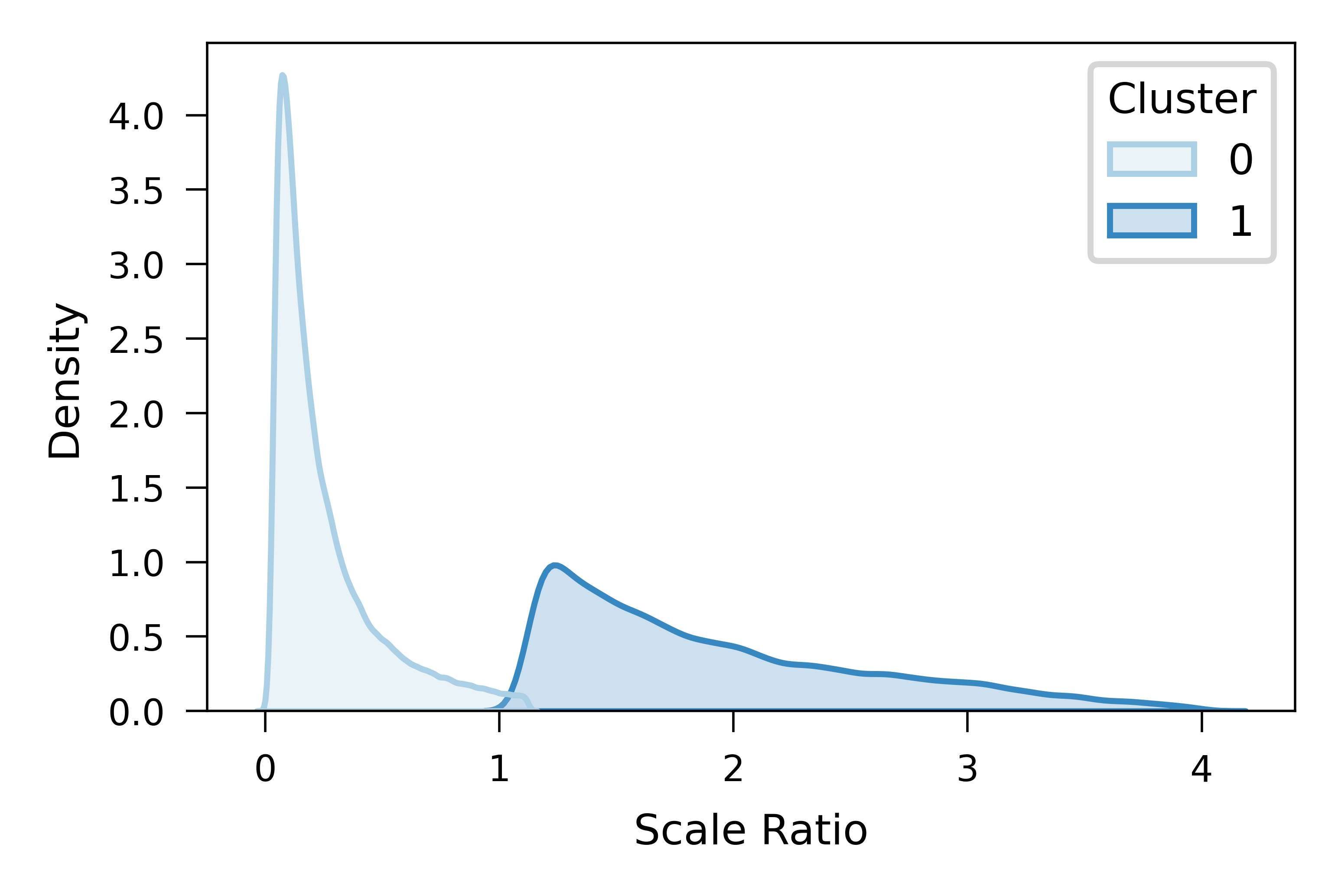}
    \caption{{Distribution of the aspect and scale ratio of the elements belonging to each cluster.}}
    \label{fig:cluster_distribution}
\end{figure}

Given the different characteristics of objects depending on their location in the image, the objective is to find a better anchor generation methodology that accounts for the perspective projection. In contrast to the default uniform generation method, the proposal is to divide the image into several key regions and obtain the best configuration for each of them independently. In order to find the optimal division in regions, a clustering analysis is performed with respect to the aspect and scale ratio of all 10 million objects in the dataset. When the K-Means algorithm is applied, the best division is obtained with two clusters after analyzing internal validity indices such as the Silhouette index. Figure \ref{fig:cluster_distribution} shows the distribution of elements in the cluster according to the values of both features. As can be seen, while the aspect ratio is similar for the elements in both clusters, there is a significant difference in the scale ratio. { The distribution plot shows that the K-Means clustering algorithm assigns to cluster 1 the majority of the larger scale elements (those that have a scale ratio greater than 1), and to cluster 0 the smaller scale objects (those that have a scale ratio below 1).}


\begin{figure}[H]
    \begin{subfigure}{\textwidth}
    \centering
    \includegraphics[width=\textwidth]{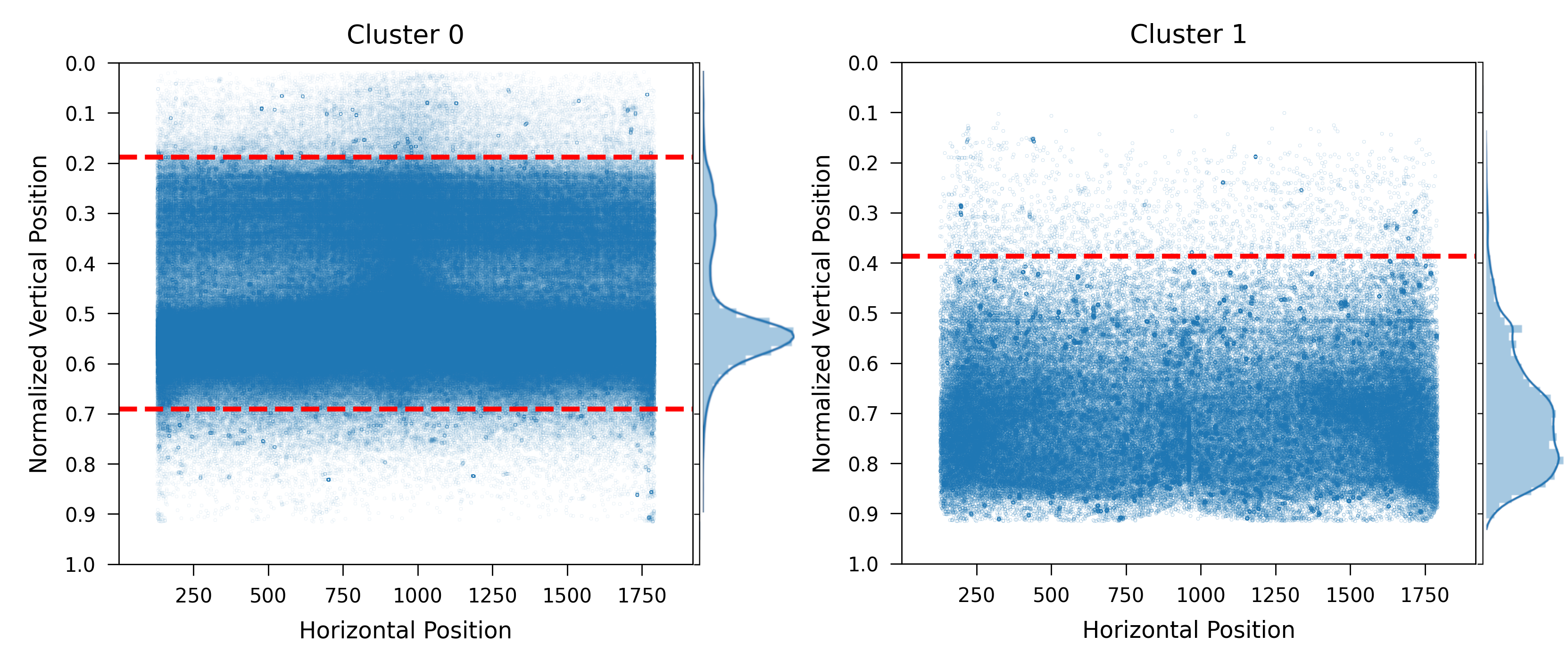}
    \caption{Distribution of the vertical and horizontal position in the image of the objects belonging to each cluster. The dotted red lines delimit the region in which there is a significant presence of objects for each cluster.}
    \label{fig:cluster_regions}
    \end{subfigure}
    \par\bigskip
    \begin{subfigure}{\textwidth}
    \centering
    \includegraphics[width=0.49\textwidth]{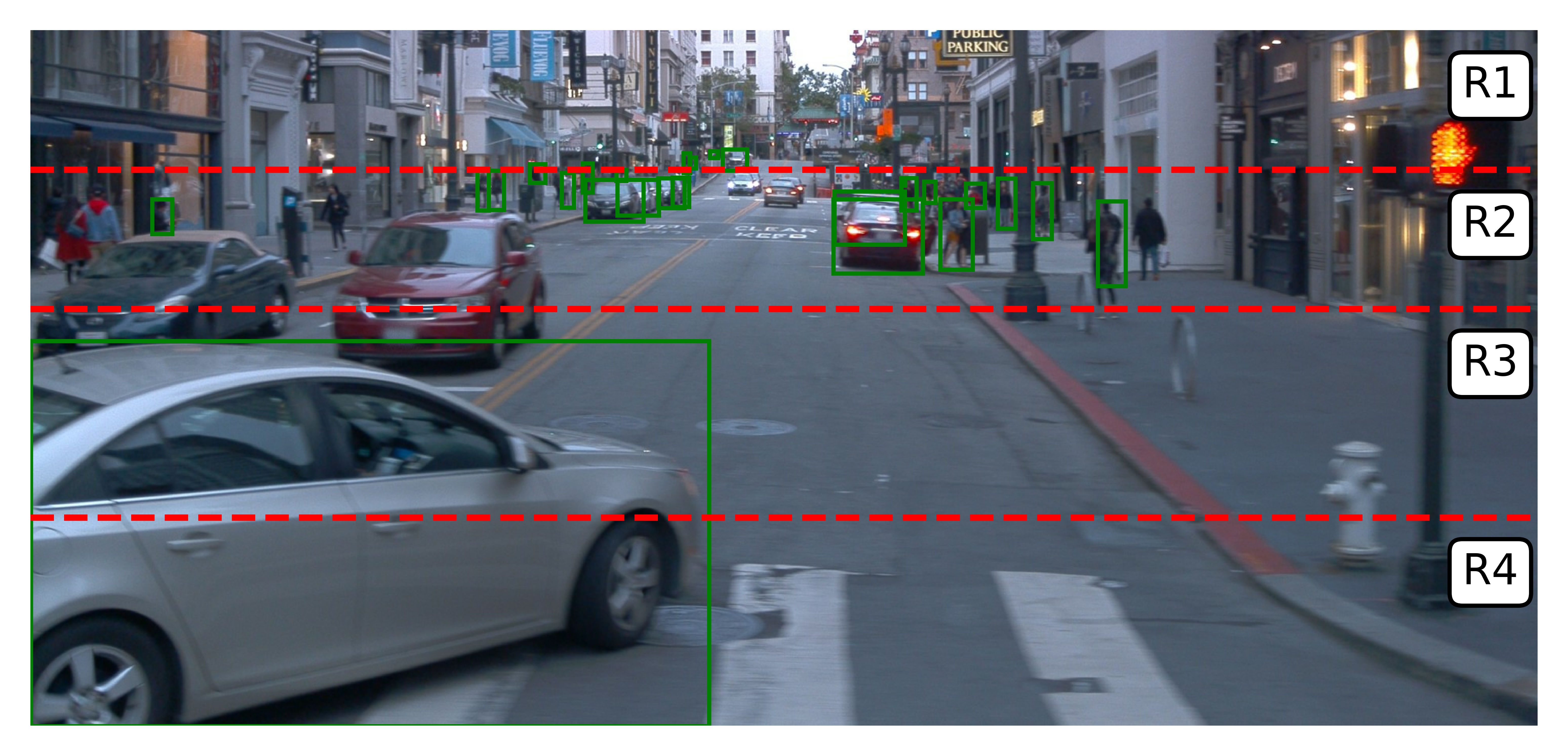}
    \includegraphics[width=0.49\textwidth]{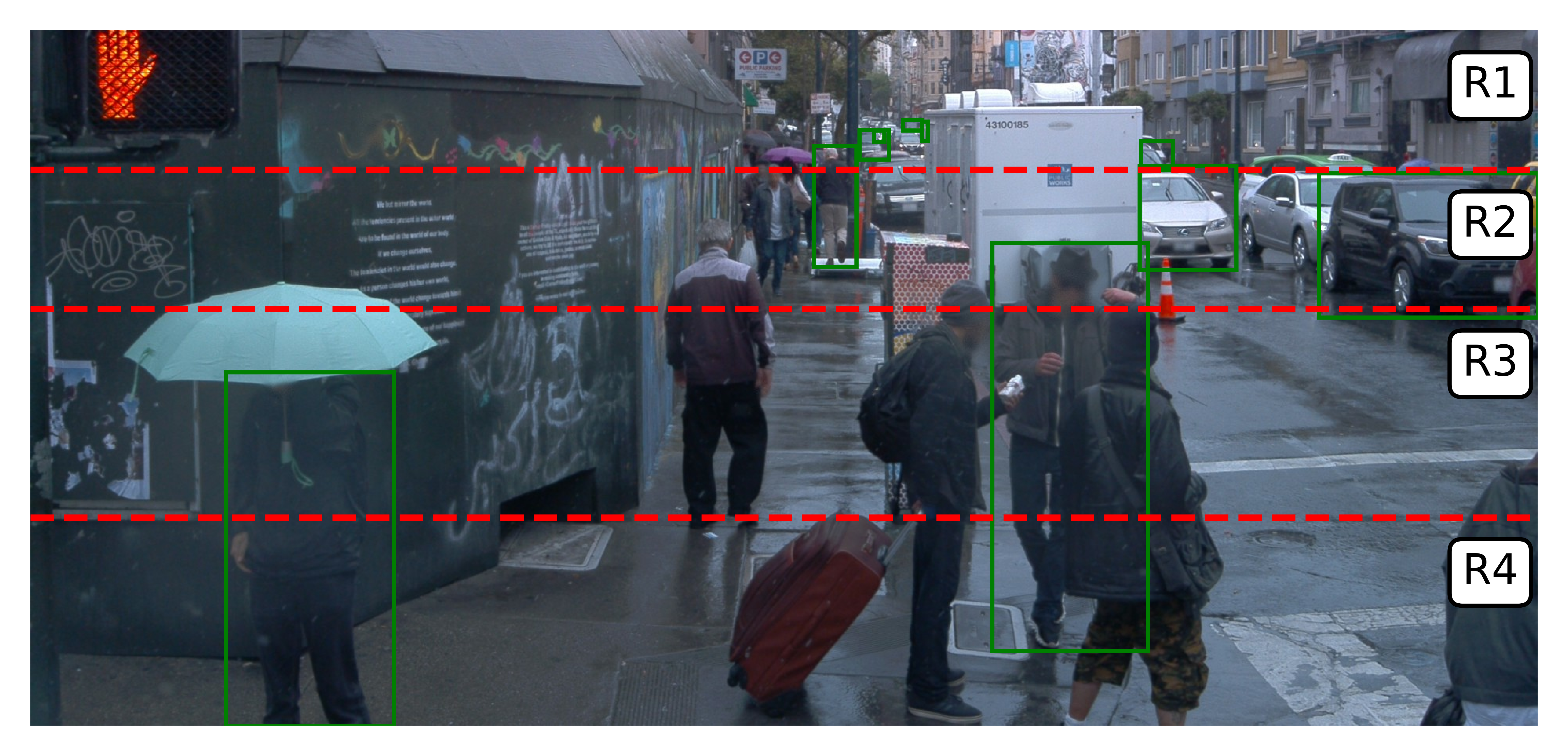}
    \caption{Example images displaying the bounds for each region. Only some labels are displayed for a clearer visualization}
    \label{fig:region_division}
    \end{subfigure}
    \caption{Division of the camera images into four regions according to the clustering study.}
    \label{fig:clust_region_division}

\end{figure}

With the clustering results, the region division can be obtained by analyzing the position in the image of the objects belonging to each cluster, which is illustrated in Figure \ref{fig:cluster_regions}. This figure plots the center of objects in both x and y-axis (horizontal and vertical position). Frontal and lateral cameras are combined for this clustering study, hence a normalized value is used for the vertical position of the objects. As can be seen in the figure, the majority of elements of cluster 1 (larger objects) are in the bottom part of the image. In contrast, elements in cluster 0 are spread across the middle and upper part of the image. {The spatial bounds that delimit the clusters can be found with the help of the density distribution of the position of elements. For each cluster, we define the interval delimited by two values $\alpha$ and $\beta$ that contains 99\% of the elements, in order to allow for the presence of outliers. Those two values define the interval in which the majority of elements of each cluster are positioned in the image. Figure} \ref{fig:cluster_regions} {displays the $\alpha$ and $\beta$ values with dotted red lines. For cluster 0, the bounds are at positions 0.188 and 0.691. For cluster 1, only the first bound (0.392) is displayed since the elements spread until the bottom of the image. If those bounds are combined, we can establish four key regions in the image that will have objects with significantly different characteristics.} The final result of the region division study is depicted in Figure \ref{fig:region_division}, in which some example images are shown with the bounds for the four regions (R1, R2, R3, and R4). These images clearly illustrate the difference in scale between objects of different regions.

\subsubsection{Evolutionary algorithm to optimize anchors per region}
Once the region division has been defined, an evolutionary algorithm (EA) is designed to find the optimal values for scale and aspect ratios for each of them. An evolutionary algorithm minimizes a fitness function by exploring a population of possible solutions over a specific number of generations. The population is formed by individuals that are represented as a chromosome, which are a combination of genes. For this problem, the chromosome is encoded as a set of 7 floating-point numbers representing the parameters that define an anchor. The first three numbers correspond to the aspect ratios and the remaining four correspond to the scale ratios. The objective is to search for the best value of each gene within defined boundaries and with a specified decimal precision. The EA creates a random initial population of a specific size. From the initial population, the algorithm combines the best individuals using single-point crossover and creates a new generation. The crossover is applied separately to the aspect ratio and scale ratio genes, so that there is no interference between the two parameters. Additionally, a low mutation probability is added to maintain genetic diversity over generations. The complete parameter configuration of the proposed EA is presented in Table \ref{tab:ga-params}. The gene boundaries are defined considering the size of images and the limitations of using ResNet-101 as a backbone network. Given the base anchor scale of 256, the specified boundaries imply that the maximum possible scale is 1024 pixels and the minimum is 16 pixels (which is the receptive field unit of the network).

\begin{table}[H]
\centering
\caption{Parameter configuration of the evolutionary algorithm}
\begin{tabular}{cc}
\hline
\textbf{Parameter} & \textbf{Value} \\ \hline
Gene boundaries & [0.06, 4] \\
Gene precision & $10^{-3}$ \\
Crossover probability & 0.8 \\
Mutation probability & 0.2 \\
Population size & 100 \\
Num. Generations & 50 \\ \hline
\end{tabular}
\label{tab:ga-params}
\end{table}

{An individual can be represented as $X = (A,S)$,  where $A = (a_1,a_2,a_3)$ defines the three possible aspect ratios and $S = (s_1,s_2,s_3,s_4)$ defines the four possible scale ratios. The decoded individual defines a set of 12 anchor configurations $B_{a_{ij}}$, which are conformed by the cartesian product between aspect and scale ratio ($B_{a_{ij}} = A \otimes S$). Therefore, the whole decoding process of an individual $X$ can be expressed as follows:}


\begin{equation}
    X = \{B_{a_{ij}}\}, \quad \text{where} \quad B_{a_{ij}} =(a_i,s_j) \quad \text{for }  i \in \{1,..,3\}, j \in \{1,..,4\}
\end{equation}

The goal of the proposed EA is to maximize the intersection between the anchors generated from the decoded individual and the ground truth boxes in the images. This is measured using the Intersection over Union (IoU) metric, which is presented in Equation \ref{eq:iou}. The IoU is defined as the ratio of area-of-overlap to area-of-union of a ground truth bounding box $B_{gt}$ and a proposed base anchor $B_a$:

\begin{equation}
    \label{eq:iou}
    \begin{aligned}
    IoU(B_{gt}, B_{a}) = \frac{area(B_{gt}\cap B_{a})}{area(B_{gt}\cup B_{a})} = \frac{area(B_{gt}\cap B_{a})}{area(B_{gt}) + area(B_{a}) - area(B_{gt}\cap B_{a})}
    \end{aligned}
\end{equation}

{The IoU metric is used for defining the EA fitness function (Eq.} \ref{eq:fitness-AG}),{
which evaluates the quality of the solutions provided by individuals in the population. For an individual $X$, the decoded chromosome generates the 12 possible anchors (4 scales and 3 aspect ratios). For each ground truth bounding box, the fitness function finds the maximum overlap obtained with the proposed anchor boxes and averages the result among all objects.} A logarithmic factor is applied in order to ensure that there are less ground truth boxes with a maximum IoU lower than 0.5 \cite{zlocha:2019}. In the equation, $K$ refers to the number of ground truth objects.


\begin{equation}
    \label{eq:fitness-AG}
    \begin{gathered}
    Fitness (X) = \dfrac{1}{K}\sum_{k=1}^K \bigg(  - \left( 1 - maxIoU(B_{gt_{k}}, X)\right) ^ 2 * \log \left(maxIoU(B_{gt_{k}}, X)\right) \bigg) \\
    maxIoU(B_{gt_{k}}, X) = \max_{\substack{i=1..3 \\ j=1..4}} \left(IoU(B_{gt_{k}}, B_{a_{ij}} ) \right)
    \end{gathered}
\end{equation}

\begin{figure}[H]
    \centering
    \includegraphics[width=\textwidth]{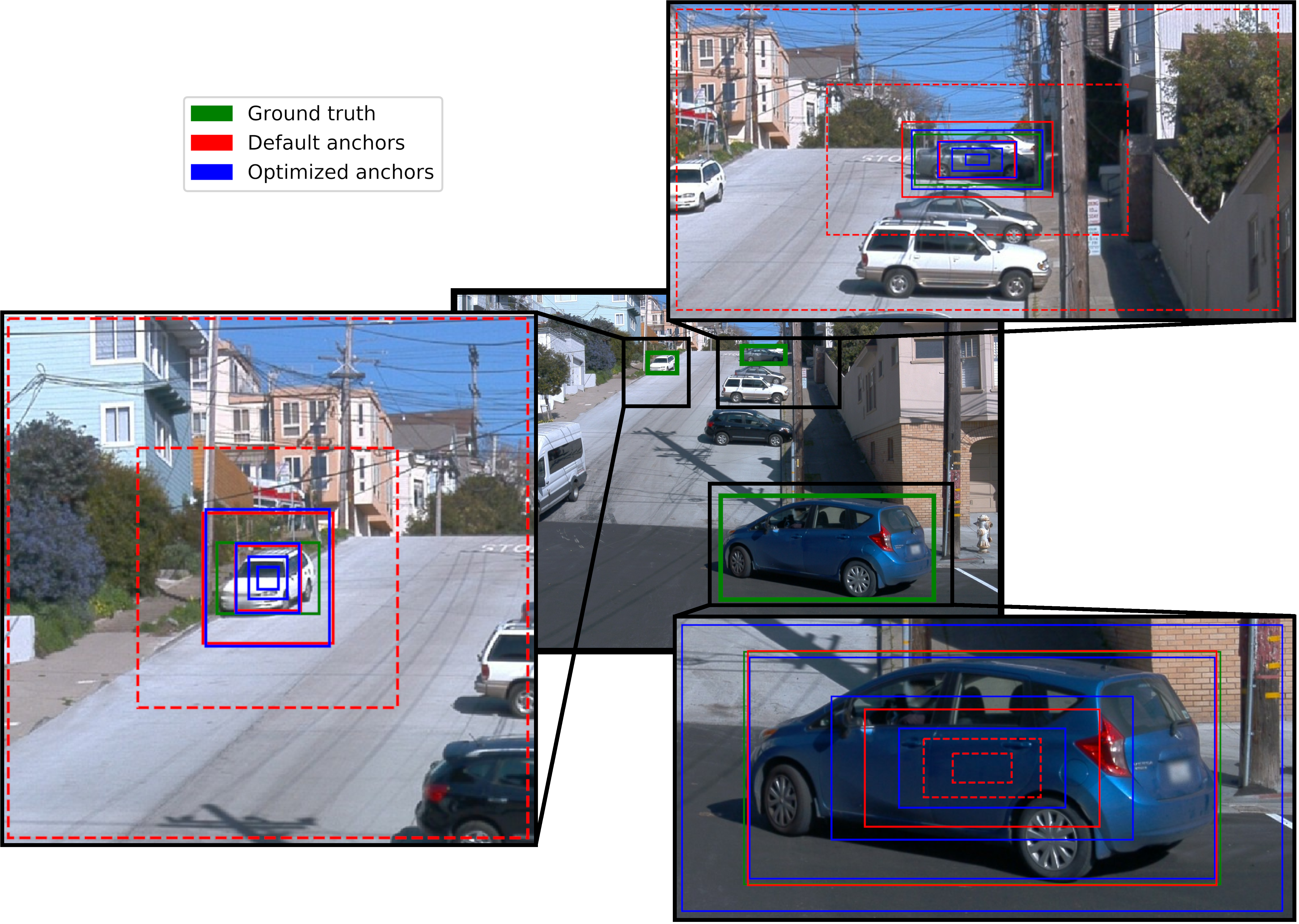}
    \caption{Difference between the uniform anchor generation procedure and our anchor optimization proposal. The complete image is displayed in the center, with three zoomed areas. The dotted red lines illustrate cases with an important anchor size mismatch when using the default configuration.}
    \label{fig:opt-anchors-example}
\end{figure}

Table \ref{tab:optimized-params-ga} presents the optimized parameters that are obtained for each region. As can be seen, the optimal values found by the evolutionary algorithm are significantly different from the original configuration. Regions at the top of the image with smaller objects have lower scale ratios. The scale in R1 and R2 ranges from values close to zero and up to a maximum of 0.5, which means that default anchors of scale 1 and 2 are not suitable. In contrast, R4 accounts for the presence of larger objects with values ranging from 0.6 until 2.9. Figure \ref{fig:opt-anchors-example} presents an example that shows the difference between the uniform anchor generation (red boxes) and our optimized proposal (blue boxes).  The dotted red boxes illustrate useless default anchors that have a great size mismatch with objects in different regions. In regions with small objects, the original strategy generates useless large-scale anchors. Analogously, in the bottom part of the image where there are objects closer to the camera, it generates very small inefficient anchors. Our proposal generates anchors with a higher matching precision, which results in a more effective detector. In the experimental study, the parameter optimization process is also analyzed separating frontal and lateral cameras, in order to check if there is a further improvement.

\begin{table}[H]
\centering
\caption{Optimized anchor configuration found with the evolutionary algorithm. The values for scale and aspect ratios for each region are provided. }
\label{tab:optimized-params-ga}
\begin{tabular}{ccc}
\hline
\textbf{Parameter} & \textbf{Region} & \textbf{Values} \\ \hline
\multirow{5}{*}{\textbf{Scale Ratio}} & Original & 0.25, 0.5, 1, 2 \\
 & R1 & 0.074, 0.158, 0.250, 0.414 \\
 & R2 & 0.082, 0.155, 0.254, 0.500 \\
 & R3 & 0.095, 0.189, 0.518, 1.557 \\
 & R4 & 0.598, 1.101, 1.800, 2.852 \\ \hline
\multirow{5}{*}{\textbf{Aspect ratio}} & Original & 0.5, 1, 2 \\
 & R1 & 0.344, 0.672, 1.801 \\
 & R2 & 0.461, 0.853, 2.246 \\
 & R3 & 0.473, 0.905, 2.497 \\
 & R4 & 0.314, 0.805, 2.136 \\ \hline
\end{tabular}
\end{table}

\subsection{ROIs spatial features concatenation}
\label{spatialfeatures}

As it was seen in the previous Section \ref{anchor-opt}, the spatial location of objects in this context plays an important role in the detection problem. Considering this fact, we propose a slight yet effective modification to the second stage of the Faster R-CNN framework. The information used by the second stage only comes from the convolutional features cropped from an intermediate feature map of the backbone network, which are then max-pooled to a $7\times7\times1024$ map. This implies that the spatial characteristics of each box proposal are lost in the ROI pooling process. The size (width and height) and position of the proposed bounding box with respect to the complete image are not taken into account by the Fast-RCNN header network. In this multi-class object detection problem with images from on-board cameras of a vehicle, the position of the box has a significant correlation with its dimensions. Therefore, including this information in the class-specific prediction and box refinement done in the second stage can enhance the localization accuracy. 

\begin{figure}[H]
    \centering
    \includegraphics{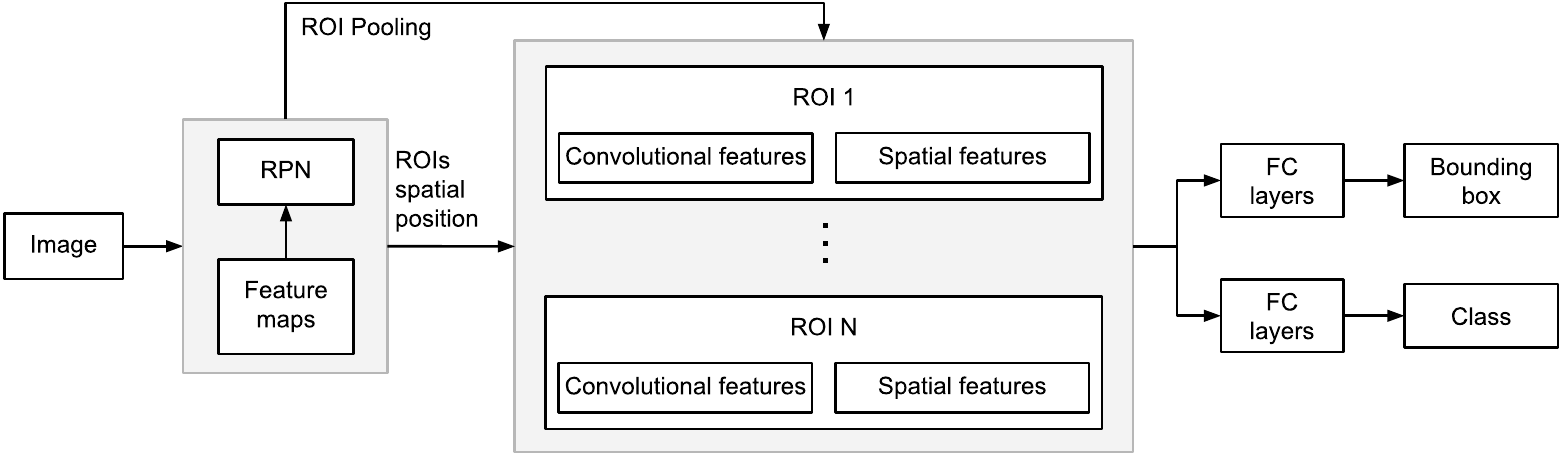}
    \caption{{Flowchart illustrating the ROIs spatial features concatenation process}}
    \label{fig:flowchart}
\end{figure}

For these reasons, our proposal is to construct those spatial features and concatenate them to the convolutional features of each ROI obtained from the cropped map. {This process is illustrated in the flowchart displayed in Figure} \ref{fig:flowchart}, and in Figure \ref{fig:fasterrcnn} with more detail. For each of the $N$ selected ROIs, cropped features are fed to the fourth block of the ResNet backbone. Then, the network is divided into two branches for different tasks: classification and bounding-box regression. In each branch, the convolutional features are flattened using a spatial average pooling, which converts the $N\times7\times7\times2048$ maps into $N\times2048$ neurons. At this point,  four features are concatenated for each ROI proposal in both branches: the width of the box, the height of the box, the horizontal position of the center of the box (X center), and the vertical position of the center of the box (Y center). This results in a layer of $N\times2052$ neurons. Those combined features are then used to predict the class and refine the boxes by means of a fully connected (FC) layer.

\subsection{Learning strategies to address class imbalance}
\label{learningstrategy}

Class imbalance is an important issue that may severely degrade the performance of detectors if it is not properly addressed \cite{Oksuz:2020:imbalance}. In object detection, there are two different class imbalance problems: background-foreground and foreground-foreground. The background-foreground imbalance is associated with the small number of positive examples (bounding boxes matching an object) compared to the number of negatives boxes (background of the image). The foreground-foreground problem refers to the imbalance between object classes of a dataset, which often leads to overfitting on the over-represented class. Since the foreground-background imbalance is unavoidable and does not depend on the specific dataset, it has attracted more interest in the recent literature. However, in the self-driving dataset used in this study, there is a significant imbalance between vehicles (80\%) and the other type of objects (pedestrians 20\% and cyclists less than 1\%). 

In order to alleviate the foreground-background imbalance, Faster R-CNN uses hard sampling techniques based on heuristic methods. In the first stage, random sampling is used for training the RPN. To avoid a bias in the learning towards negative samples, 128 positive and 128 negative examples are randomly selected to contribute to the loss function. In the second stage, the employed strategy is to limit the search space. For the Fast R-CNN detection network, only the best $N$ ROIs are selected according to their objectness scores, while maintaining a 1:3 positive-negative ratio. However, this methodology presents the problem that all examples are equally weighted once they are sampled. The original Faster R-CNN implementation does not take into account the presence of under-represented objects, which limits performance under high imbalance. Therefore, we aim to design a better training procedure that addresses the foreground-foreground imbalance in this scenario. 

Our proposal is to study how the loss function of two-stage detectors can be modified using cost-sensitive re-weighting techniques, which has been one of the main approaches in the literature for class imbalance problems \cite{Cui:2019}. These techniques are based on assigning relatively higher costs to minority instances, hence building a better class-balanced loss. In this work, we explore two re-weighting alternatives that can improve overall detection accuracy: balance the loss function with weights that are proportional to the class distribution; and the use of focal loss, which has been traditionally used for one-stage detectors.

\subsubsection{Weight assignment based on class distribution}

The first approach is to assign different weights to instances of each class depending on the class frequency distribution. This simple yet effective method has been widely used in many computer vision tasks \cite{Huang:2016}. As stated before, in Faster R-CNN, the training is framed as a multi-task learning problem that combines classification and bounding box regression. We propose to modify the loss function by adding a weight $w_i$ parameter to the classification and regression terms. The complete loss function with weights assignment is expressed in Equation \ref{weights}:

\begin{equation}
\label{weights}
\begin{gathered}
    L({p_i},{t_i}) = L_{cls} + L_{reg} =
    \dfrac{1}{N_{cls}} \sum_{i} w_i L_{cls}(p_i, p_i^*)
    +  \dfrac{\lambda}{N_{reg}} \sum_{i} w_i p_i^*  L_{1}^{smooth} (t_i - t_i^*) \\ 
    L_{cls}(p_i,p_i^*) = CE(p_i,p_i^*) = -p_i^* \log p_i - ( 1- p_i^*) \log (1 - p_i)
\end{gathered}
\end{equation}

where $p_i$ is the predicted probability of proposal $i$ being an object, and $p_i*$ is the groundtruth label (0 or 1). $t_i$ are the four predicted box coordinates, and $t_i*$ are the ground truth coordinates.  $w_i$ is the weight assigned depending on the ground truth class. $N_{cls}$  and  $N_{reg}$ are normalization terms which are set to the RPN mini-batch size, which is typically 256. $L_{cls}$ is the binary cross-entropy function and $L_{reg}$ is the smooth L1 loss function. 

Note that we have described only the RPN loss function for simplicity, but the modification is done in both stages. The loss function in the second stage is the same, except that in the RPN the classification problem is binary (background vs object) while in the Fast R-CNN header network the classification is multi-class. This implies that the multi-class cross-entropy function is used. In this case, the normalization parameters are set to the number of ROI proposals that pass to the second stage.

{In order to obtain the best possible performance, a grid search is performed to find the optimal weight value for each class. The search involves experimenting with different sets of weights, considering four possible values $(0.3,0.5,0.7,0.9)$ that can be assigned to vehicles and pedestrians. The maximum weight $(1.0)$ is kept for cyclists since they are extremely underrepresented.}

\subsubsection{Reduced Focal Loss}

The second re-weighting alternative that is explored is focal loss, which was introduced in \cite{Lin:2020:Focal} as an improvement to the RetinaNet one-stage detector. The focal loss function belongs to another line of work that assigns weights according to sample difficulty. These methods assume that training a detector with hard examples improves performance. A method of this family that has been used in two-stage detectors is Online Hard Example Mining (OHEM) \cite{Shrivastava:2016}. However, OHEM requires more memory consumption and increases training time. Although focal loss was found to be more effective than OHEM for imbalance problems, it has not been used extensively in two-stage detectors. In this work, we study the application of the focal loss function to the Faster R-CNN framework and propose a modification to better adjust it to its characteristics.

Focal loss (FL) assigns higher weights to hard examples, with the aim of alleviating the high background-foreground imbalance. As it shown in Equation \ref{focalloss}, focal loss modifies the standard cross-entropy equation by adding a factor  $(1 - p_i)^\gamma$. When $\gamma > 0$, the relative loss for easy and well-classified samples is reduced, putting the effort on the classification of hard examples.

\begin{equation}
\label{focalloss}
    FL(p_i) = -\alpha(1 - p_i)^\gamma \log(p_i)
\end{equation}

Although this function was introduced to reduce the influence of easy background examples, it also has an effect on the foreground-foreground imbalance. Instances from minority classes often have higher losses since they are rare and their features are usually poorer. However, the direct application of focal loss in two-stage detectors presents several problems. Firstly, focal loss can have as a side-effect that the learning is biased towards noisy or mislabeled data, which is also hard to classify \cite{Cui:2019}. Furthermore, focal loss contradicts slightly the behaviour of two-stage detectors: RPN aims to maximize recall (allowing false positives), while the job of Fast R-CNN header network is to classify proposals correctly. An extreme focus on hard samples can reduce the recall of RPN, which implies that lower-quality proposals pass to the second stage. Therefore, inspired by the work in \cite{Sergievskiy:2019:RFL},  we propose a modified version called reduced focal loss (RFL).

Reduced focal loss aims to perform hard example mining but softening the effect of difficult samples. This is achieved by applying the factor only to instances with losses that are above a certain threshold. The loss of samples that are below the threshold remains unaltered, which means that is the same as the original cross-entropy loss. This approach is formulated in Equation \ref{RFL}. RFL is applied in the classification term of the loss function in both stages of the Faster R-CNN detector. 

\begin{equation}
\label{RFL}
\begin{gathered}
    RFL(p_i) =  - \alpha RF(p_i,th) \log(p_i)\\
    \\
    RF(p_i,th) = \begin{cases}
    1 & \text{if $p_i < th $} 
    \\
    \dfrac{(1 - p_i)^\gamma}{th^\gamma} & \text{if $p_i \geq th $}
    \end{cases}
    \end{gathered}
\end{equation}

Following \cite{Sergievskiy:2019:RFL}, the thresholds are fixed to 0.5 and 0.25 for the RPN and Fast R-CNN respectively. RFL is more suitable than the original focal loss for two-stage detectors, and can significantly improve the performance over rare and difficult instances. Another important advantage that is obtained when using focal loss here is that all anchors can be considered in the RPN loss function. The sampling process in the original RPN discards many anchor boxes that can be useful for the training process, especially those hard negatives that are close to an object.


\subsection{Ensemble model}

Ensemble models can be key to further improve the robustness of detection systems. However, ensembling object detectors is not straightforward due to the specific particularities of this problem: multiple classes with different shapes, overlapping bounding boxes, etc. \cite{Xu:2020:Ensemble}. In this work, we propose to build an ensemble that combines the output of models from the different learning strategies presented in the previous Section \ref{learningstrategy}. The ensemble model is based on Non-Maximum Suppression (NMS) \cite{Hosang:2017}, and uses the affirmative ensembling strategy proposed in \cite{Casado:2020}. With this strategy, when one of the methods proposes the presence of an object in a region, such a detection is considered as valid. Later, the NMS algorithm is applied to merge detections by removing redundant overlapping boxes. NMS selects proposals in descending order of confidence scores, and discards boxes that have an IoU overlap with already selected boxes greater than a pre-defined threshold. Our proposal is to merge the output of three models with different training schemes: original Faster R-CNN training, re-weighting proportional to class distribution, and reduced focal loss re-weighting. With this approach, the aim is to enhance detection accuracy without a significant increase in computation. If the predictions from the single models are obtained in parallel with different devices, the overhead introduced by the NMS algorithm is minimal. The only parameter that has to be tuned for the NMS ensemble is the IoU threshold. We explored values ranging from 0.5 and 0.9, and found consistent and similar results on values between 0.6 and 0.8. Therefore, the IoU threshold was fixed to the intermediate value 0.7, which is also the same value used internally in the ROI selection process of Faster R-CNN.

Furthermore, Test-Time Augmentation (TTA) is also explored in the experiments as another ensemble technique. TTA is to create random modifications of the test images, obtaining predictions with a model for each of them, and then ensemble the results. The only augmentation that is exploited is scale augmentation, with the objective of improving the detection of small objects. The test images are resized using the factors 0.8, 1, and 1.2 with respect to the original image scale. This is done for each of the three considered models, and detections are merged as explained above.

\subsection{Other implementation details}
For the implementation, the popular TensorFlow Object Detection API as been used \cite{TensorflowAPI}. Except for the proposed modifications, the original Faster-RCNN implementation is followed in terms of parameter choice. The architecture is trained end-to-end, which means that RPN and header networks are jointly trained. The original resolution of the images is maintained for frontal $(1920\times1280)$ and lateral $(1920\times886)$ cameras. The COCO pre-trained Faster R-CNN model available in the TensorFlow repository is used for initializing the network weights. Transfer learning is used since it is a common practice in the object detection field to avoid excessive training times \cite{Alvaro:2018}. For each experiment, the models are fine-tuned for $500k$ iterations, using the SGD optimizer with learning momentum 0.9. The initial learning rate is set to 0.001, and decrease by 10x after $200k$ and $400k$ steps. {For training the models, a batch size of 1 is used, which was the maximum allowed by the GPU memory given that the images have very high resolution.} The ROI proposals mini-batch size is increased to 256 during training in order to accelerate convergence \cite{Lin:2017:FPN}. At test time, the original value of 300 proposals is kept. The only data augmentation technique used in single models is random horizontal flip.

\section{Results and Discussion}
\label{section4}
This section reports and discusses the results obtained from the experiments carried out in this study. {First of all, the baseline results with the original model are presented. Later, the experiments are reported following the same order in which the methods are presented in Section} \ref{section3}. {This allows performing a comprehensive analysis of the effectiveness of all proposed modifications in an incremental manner.}

\subsection{Evaluation metric}

In this study, the Average Precision (AP) metric is used to measure performance. AP is the reference object detection metric in the literature \cite{Zou:2019}. It is calculated by tracking the interpolated precision/recall curve. As shown in Equation \ref{maxrecall}, the first step is to set the precision for recall $r$ to the maximum precision obtained for any recall $r' \geq r$. The AP is then calculated as the area under this curve by numerical integration. As can be seen in Equation \ref{integral}, this value can be approximated by the sum of the precision at every $k$ where the recall changes, multiplied by the change in recall $\Delta r(k)$.

\begin{equation}
    \label{maxrecall}
    p(r) = \max_{r': r'>r} p(r')
\end{equation}

\begin{equation}
    \label{integral}
    AP = \int_0^1 p(r)dr \approx \sum_ {k=1}^N p(k) \Delta r(k)
\end{equation}

The IoU is used to determine whether the predicted bounding boxes are considered true or false positives. As stated before, the IoU is defined as the area of overlap between a predicted box and a ground truth box divided by their area of union. A prediction is considered true positive if the IoU value is above a certain threshold, being a false positive otherwise.  Ground truths objects with no matching detections are considered false negatives. In the Waymo dataset, the IoU thresholds are defined as 0.7 for vehicles and 0.5 for pedestrians and cyclists\cite{Waymo:2019}. In the next sections that present the experimental results, the accuracy for each class as well as the average AP of the three classes are reported. {The precision values are divided into the two difficulty levels provided by Waymo (L1 and L2).}

{Furthermore, the computational efficiency of the models is analyzed since it is an essential aspect in the autonomous driving scenario. The training and inference times of the models are reported, considering a batch size of 1 image.  The input video in the Waymo data comes at 10Hz, hence 10 frames per second (FPS) can be considered real-time speed in this context. Besides speed, other metrics are analyzed such as memory usage, the number of floating-point operations (FLOPS), and the number of parameters. These metrics allow comparing the models independently from the hardware used in the experiments.}

\subsection{Baseline results}
\label{baseline}
{This section presents the results obtained with the original Faster R-CNN model and other popular detection frameworks that are implemented in the TensorFlow Object Detection API. These results, which are presented in Table} \ref{tab:baseline}, {will be the baseline for the comparison to validate our proposal.} 

Two baseline results are provided for the original Faster R-CNN model. The first baseline is the one provided by Waymo \cite{Waymo:2019} using a COCO pre-trained Faster R-CNN with ResNet-101, which is also the base of our proposal. However, Waymo does not provide in the report any details about the parameter configuration and the accuracy over cyclists is missing. Therefore, we also run our own experiment with the default Faster R-CNN provided in the Tensorflow API, in order to allow for a full and fair comparative study.  Our experiment with the default configuration obtains better results than the one reported by Waymo (2.1\% in vehicles and 2.4\% in pedestrians), hence it will be the baseline used in future comparisons. The analysis will be mainly focused on the Level 2 AP metric since it also includes all objects in Level 1.

{Furthermore, Table} \ref{tab:baseline} { compares the performance of Faster R-CNN with other one-stage detectors that have been extensively used in the recent literature such as RetinaNet, CenterNet, and YOLOv3. These one-stage models with different characteristics are also fine-tuned from the available checkpoints pre-trained on COCO. RetinaNet and YOLOv3 are both anchor-based detectors, while CenterNet is an anchor-free model. RetinaNet and CenterNet use the same ResNet-101 backbone as Faster R-CNN, but with feature pyramid networks, which increases computation cost. In contrast, YOLOv3 uses the DarkNet-53 feature extractor, which is more efficient than ResNet.} 

{As can be seen, the original Faster R-CNN outperforms the one-stage models in all the AP metrics. The difference in accuracy is more significant in the cyclist class, which indicates that one-stage models are less robust to the presence of imbalanced data. RetinaNet is the most competitive model in terms of average accuracy among the studied one-stage detectors, while CenterNet obtains a slightly higher precision on pedestrians. YOLOv3 achieves similar AP values to the other one-stage models in the pedestrian and cyclist classes. However, it suffers a significant precision drop in the vehicle class. This is due to the harder IoU threshold of this class, as YOLOv3 does not perform well for thresholds greater than 0.5} \cite{Yolov3:2018}.  

\begin{table}[H]
\centering
\caption{{Baseline results obtained with the original Faster R-CNN model and other one-stage detection frameworks.}}
\label{tab:baseline}
\begin{tabular}{ccccccccc}
\hline
\textbf{} & \multicolumn{2}{c}{\textbf{Vehicle}} & \multicolumn{2}{c}{\textbf{Pedestrian}} & \multicolumn{2}{c}{\textbf{Cyclist}} & \multicolumn{2}{c}{\textbf{Average}} \\ \hline
 \textbf{Model} & \textbf{L1} & \textbf{L2} & \textbf{L1} & \textbf{L2} & \textbf{L1} & \textbf{L2} & \textbf{L1} & \textbf{L2} \\ \hline
Waymo baseline \cite{Waymo:2019} & 63.70 & 53.30 & 55.80 & 52.70 & - & - & - & - \\
Original Faster R-CNN & \textbf{64.82} & \textbf{55.44} & \textbf{58.91} & \textbf{55.12} & \textbf{43.19} & \textbf{38.28} & \textbf{55.64} & \textbf{49.61} \\ \hline
RetinaNet & 61.21 & 50.43 & 55.59 & 52.02 & 29.09 & 24.08 & 48.63 & 42.18 \\
CenterNet & 50.70 & 43.25 & 57.23 & 53.83 & 30.55 & 23.89 & 46.16 & 40.32\\
YOLOv3 & 49.82 & 41.76 & 53.89 & 50.25 & 31.12 & 24.96 & 44.94 & 38.99
\end{tabular}
\end{table}

\subsection{Anchor optimization and spatial features}
\label{results1}
Section \ref{anchor-opt} presented the importance of the spatial properties of objects in the context of autonomous driving, and the strong correlation between object position in the image and bounding box dimension. Therefore, the first analysis that is presented is the assessment of the anchor optimization procedure and the addition of spatial features in the second stage of Faster R-CNN. For this purpose, we conduct several experiments that show the impact of the proposed modifications on the performance. 

\begin{figure}[H]
     \centering
     \includegraphics{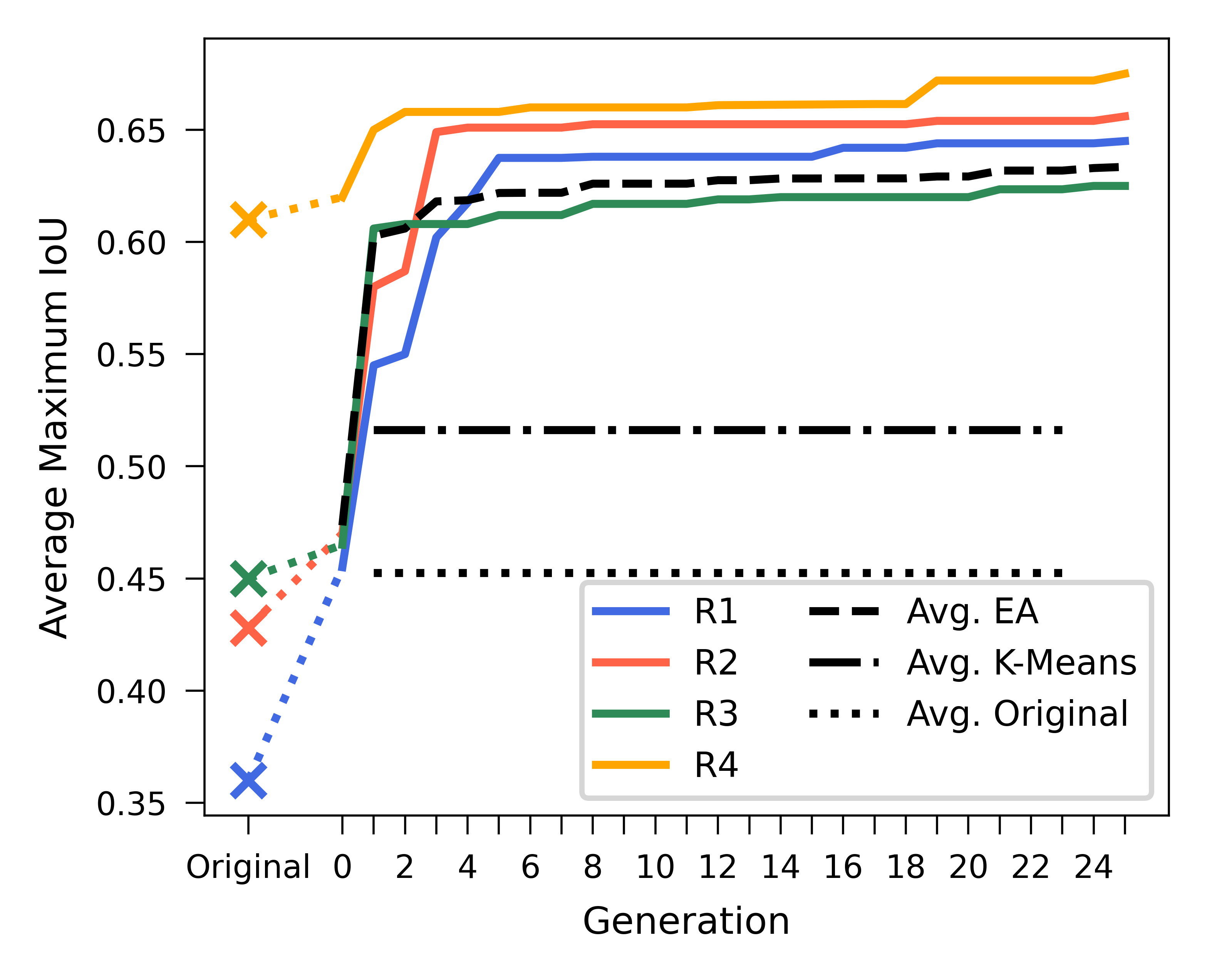}
     \caption{{Evolution of the EA used to optimize the generation of anchors. The dotted lines represent the average value of the four regions.}}
     \label{fig:genetic-algorithm}
\end{figure}

First of all, we evaluate the effect in the network initialization of the optimized anchor parameters found by the EA.  Figure \ref{fig:genetic-algorithm} illustrates the evolution over generations of the average IoU between ground truth boxes and proposed anchors in each region. As can be seen, the evolutionary algorithm obtains a significant improvement over the original parameters. The average value obtained with the default uniform anchors is around 0.45 IoU. This demonstrates that the original pre-defined scales and aspect ratios are not suitable in this context. Due to the perspective projection of cameras, the IoU is very far from optimal in regions at the top of the image that contain smaller objects (R1, R2, R3). The evolutionary algorithm converges to a much better solution for each region, obtaining an average IoU of 0.64. This indicates that there will be many more anchors that match with object boxes and that the Faster R-CNN training will be more effective. {Furthermore, Figure} \ref{fig:genetic-algorithm} {also displays the result obtained if the anchors are optimized using K-means clustering as it is done in the YOLO detector} \cite{Yolov3:2018}. {This simpler approach obtains an average IoU of 0.516, which shows that the proposed EA with the custom fitness function provides a better solution to the problem of anchor optimization in this context.}

The results of the experiments conducted in this section are reported in Table \ref{tab:res1}. This table shows the improvement obtained over the original Faster R-CNN model, which was provided in Table \ref{tab:baseline}. The first experiment performed is to concatenate the ROIs spatial features in the Fast R-CNN header network. This allows to separately evaluate the importance of this novel proposal. We maintain the original anchor configuration and only modify the second stage of the Faster R-CNN by adding the spatial features module. This modification already obtains an average 0.85\% AP improvement, that is more important in pedestrians with a 1.95\% increase. This finding confirms the fact that including spatial properties from the ROIs proposed by the RPN can improve localization accuracy. Although the cropped convolutional features contain rich information, the position of objects in the images plays a key role due to the perspective projection of cameras. Therefore, having the spatial information of the proposals for the final classification and box-refinement seems essential in this context.

Afterwards, the spatial features module is combined with the per-region anchor optimization procedure. Two experiments are performed: optimizing the base anchors for the complete dataset, and separating the images from frontal and lateral cameras. Since they capture images with different dimensions, better results may be obtained if the evolutionary algorithm is applied independently in both subsets. When employed over the whole dataset, our anchor optimization method boosts the performance by more than 5.4\% and 4.1\% average AP with respect to the baseline in L1 and L2 respectively. Again, the greatest improvement is obtained in pedestrian detection with more than a 9\% increase. These results demonstrate that the uniform anchor configuration of the default Faster R-CNN is not suitable for this scenario. The different size of the objects in different regions of the images requires a better anchor generation procedure. The per-region optimization proposal better matches the shape of objects, allowing a more effective training. Furthermore, when searching for the optimal base anchors by separating frontal and lateral cameras, an additional enhancement is obtained. In summary, the model with optimized anchors obtains an average AP of 54.37, which is a significant enhancement compared to the original 49.61. In the experiments presented in the next section (the study of different learning strategies and the ensemble models), this model is used as the base: Faster R-CNN with spatial features concatenation and anchor optimization separating frontal and lateral cameras. These improvements can be easily extended to other anchor-based detection frameworks since our proposal does not rely on this specific implementation.

\begin{table}[H]
\centering
\caption{AP values obtained using the spatial features concatenation model and anchor optimization. The baseline result from Table \ref{tab:baseline} obtained with the original model is provided for comparison. F/L refers to optimizing anchors for frontal and lateral camera separately.}
\label{tab:res1}
\begin{tabular}{ccccccccc}
\hline
 & \multicolumn{2}{c}{\textbf{Vehicle}} & \multicolumn{2}{c}{\textbf{Pedestrian}} & \multicolumn{2}{c}{\textbf{Cyclist}} & \multicolumn{2}{c}{\textbf{Average}} \\ \hline
\textbf{Model} & \textbf{L1} & \textbf{L2} & \textbf{L1} & \textbf{L2} & \textbf{L1} & \textbf{L2} & \textbf{L1} & \textbf{L2} \\ \hline
Original Faster R-CNN & 64.82 & 55.44 & 58.91 & 55.12 & 43.19 & 38.28 & 55.64 & 49.61 \\
ROIs spatial features & 65.91 & 55.56 & 61.30 & 57.07 & 44.16 & 38.76 & 57.12 & 50.46 \\
Anchor optimization & 68.19 & 56.33 & 66.73 & 64.48 & 48.21 & 40.57 & 61.04 & 53.79 \\
F/L Anchor optimization & \textbf{68.51} & \textbf{57.29} & \textbf{66.94} & \textbf{64.68} & \textbf{49.07} & \textbf{41.16} & \textbf{61.50} & \textbf{54.37} \\ \hline
\end{tabular}
\end{table}

\subsection{Learning strategies to address class imbalance}

The next step of the experimental study is to evaluate the different training methodologies that have been proposed to address the foreground-foreground class imbalance. As it was explained in Section \ref{dataset}, in this problem there is a high imbalance between vehicles and pedestrians. Furthermore, the number of cyclist boxes is very reduced, which explains the performance drop in this class seen in Table \ref{tab:baseline}. Table \ref{tab:res2} presents the results obtained with the three different learning strategies: the original Faster R-CNN training, the re-weighting according to class distribution (which is noted as class weights), and re-weighting according to sample difficulty (focal loss). Note that all these models already include the modifications studied in Section \ref{results1}, that is, the anchor optimization and the spatial features concatenation.

For the re-weighting proportional to class distribution, a grid search with different sets of weights is performed. In Table \ref{tab:res1}, the two configurations that obtained better average performance are displayed: $(0.5,0.9, 1.0)$ and $(0.7, 0.9, 1.0)$ for weights of vehicles, pedestrians, and cyclists respectively. As can be seen, with this approach the accuracy of minority instances improves, especially over cyclists. In the first experiment $(0.5, 0.9, 1)$, in which the weights of vehicles is reduced more aggressively, a 1.57\% increase in cyclists AP is obtained. In the second case $(0.7, 0.9, 1)$, the improvement is a smaller 1.05\% increase. However, the AP over pedestrians remains with similar values compared to the original training, or even slightly worse in the latter case. Furthermore, the improvement over cyclists comes with the cost of an undesired performance decrease in vehicles. These results suggest that this alternative is not the best option if it were to be used as a single model. However, it seems a suitable alternative if the main focus is to improve on classes with an extremely small number of examples. In this dataset, cyclist labels represent less than 1\% of the total number of instances. Therefore, this re-weighting methodology is effective to improve accuracy over this class. In order to account for all involved classes, a better solution will be to combine several complementary models with different sets of weights.

The other alternative that is considered is to re-weight according to sample difficulty. For this case, two experiments are performed: the standard focal loss as defined in \cite{Lin:2020:Focal} and the reduced focal loss version. As expected, the original focal loss does not provide a significant enhancement in performance.  In contrast, the reduced version is found to be far more effective. Reduced focal loss obtains an average 1.37\% improvement over the default training scheme. It is important to mention that the greatest AP increase is achieved over pedestrians, which is the minority class. However, the improvement over the cyclist class is not as important as the one obtained with the class weights alternative. This suggests that the number of instances is not sufficient for focal loss to have an effect. In general, these results show that focal loss can be used effectively in two-stage detectors if it is properly adjusted with this simple modification. The reduced focal loss training provides the best single model that has been found in the experiments, obtaining a 6.13\% AP increase compared to the original Faster R-CNN baseline.

\begin{table}[H]
\centering
\caption{AP values obtained using the different learning strategies to address class imbalance. The best model of Table \ref{tab:res1} uses the original Faster R-CNN training procedure. In class weights models, the numbers represent the weights assigned to vehicles, pedestrians and cyclists respectively.}
\label{tab:res2}
\begin{tabular}{ccccccccc}
\hline
 & \multicolumn{2}{c}{\textbf{Vehicle}} & \multicolumn{2}{c}{\textbf{Pedestrian}} & \multicolumn{2}{c}{\textbf{Cyclist}} & \multicolumn{2}{c}{\textbf{Average}} \\ \hline
\textbf{Model} & \textbf{L1} & \textbf{L2} & \textbf{L1} & \textbf{L2} & \textbf{L1} & \textbf{L2} & \textbf{L1} & \textbf{L2} \\ \hline
Best model Table \ref{tab:res1} & 68.51 & 57.29 & 66.94 & 64.68 & 49.07 & 41.16 & 61.50 & 54.37 \\ \hline
Class weights (0.5, 0.9, 1) & 66.56 & 56.22 & 67.04 & 64.81  & \textbf{50.21} & \textbf{42.73} & 61.27 & 54.58 \\

Class weights (0.7, 0.9, 1) & 66.85 & 56.58 & 66.59 & 63.48 & 49.56 & 42.21 & 61.00 & 54.09 \\ \hline

Focal loss & 68.73 & 57.15 & 68.14 & 65.45 & 47.79 & 41.54 & 61.55 & 54.71 \\ 
Reduced focal loss & \textbf{69.37} & \textbf{58.55} & \textbf{69.57} & \textbf{66.95} & 49.13 & 41.73 & \textbf{62.69} & \textbf{55.74} \\ \hline
\end{tabular}
\end{table}

\subsection{Ensemble models}

The previous section demonstrated that the considered learning strategies have different strengths and perform better over different classes. Therefore, we propose to build an ensemble model fusing the output of three models to enhance the detection performance. The models with the best average AP for each learning strategy in Table \ref{tab:res2} are selected: the original training, the class weights with weights set (0.5, 0.7, 0.9), and the reduced focal loss. With these models, two different ensemble techniques have been tested: the NMS ensemble and its combination with test-time augmentation. The results are presented in Table \ref{tab:res3}, in which they are compared to the best single model that was obtained using reduced focal loss.

As can be seen, the improvement obtained in this case is very significant. While it is difficult to build a single model with a strong performance on all classes, ensemble models provide a solution to increase the robustness of predictions. Compared to the best single model, the NMS ensemble achieves a 2.47\% AP increase. The improvement is consistent for all classes and is even more important for both minority classes. Moreover, the use of TTA further enhances localization accuracy over vehicles and pedestrians. The NMS ensemble with TTA was the best model that was obtained at Waymo's online challenge, providing an average 9.69\% AP increase over the initial Faster R-CNN baseline.

\begin{table}[H]
\centering
\caption{AP values obtained with the ensemble models. The best model of Table \ref{tab:res2} refers to the reduced focal loss model.}
\label{tab:res3}
\begin{tabular}{ccccccccc}
\hline
 & \multicolumn{2}{c}{\textbf{Vehicle}} & \multicolumn{2}{c}{\textbf{Pedestrian}} & \multicolumn{2}{c}{\textbf{Cyclist}} & \multicolumn{2}{c}{\textbf{Average}} \\ \hline
\textbf{Model} & \textbf{L1} & \textbf{L2} & \textbf{L1} & \textbf{L2} & \textbf{L1} & \textbf{L2} & \textbf{L1} & \textbf{L2} \\ \hline
Best model Table \ref{tab:res2} & 69.37 & 58.55 & 69.57 & 66.95 & 49.13 & 41.73 & 62.69 & 55.74 \\
NMS Ensemble & 69.61 & 59.69 & 72.81 & 69.28 & 52.27 & 45.67 & 64.90 & 58.21 \\
NMS Ensemble + TTA & \textbf{70.76} & \textbf{61.37} & \textbf{74.17} & \textbf{70.88} & \textbf{52.31} & \textbf{45.70} & \textbf{65.75} & \textbf{59.30} \\ \hline
\end{tabular}
\end{table}

\begin{figure}[H]
    \centering
    \includegraphics[width=\textwidth]{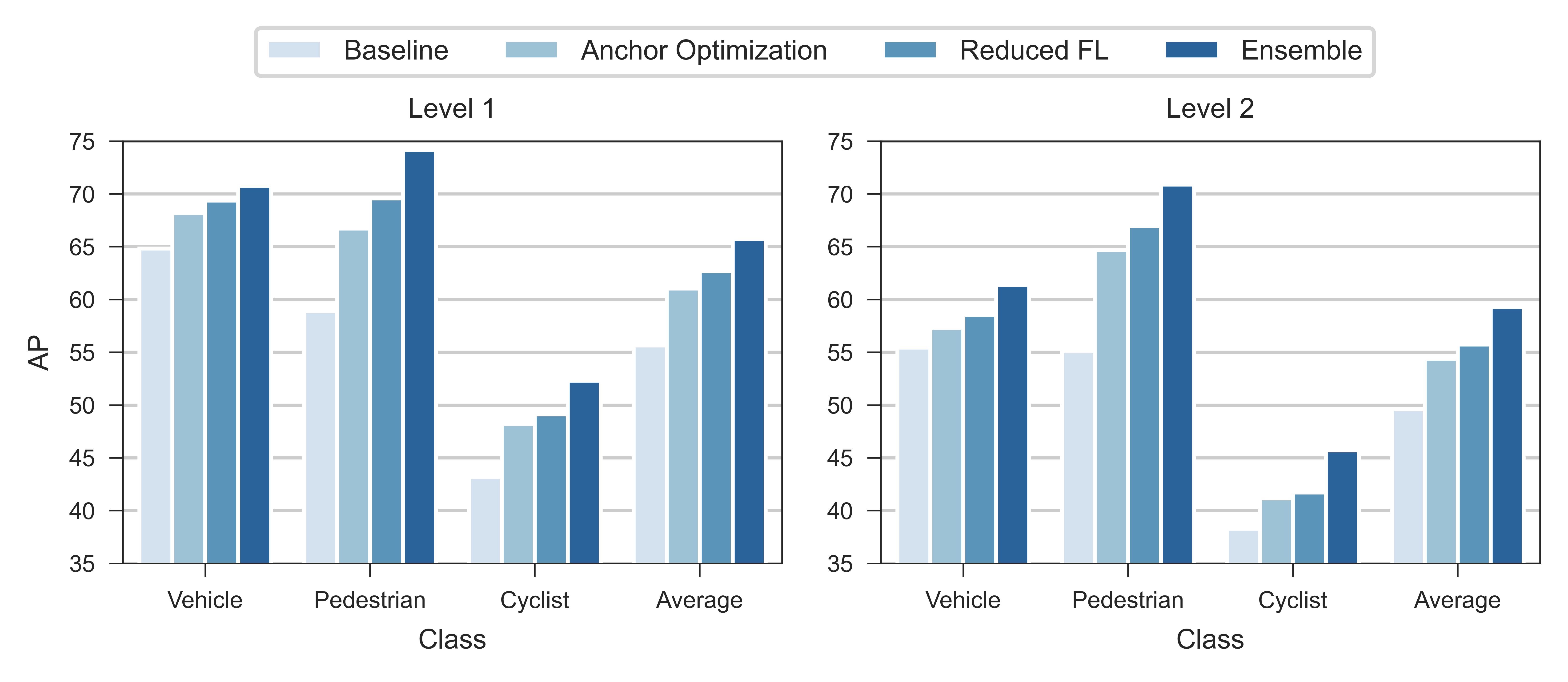}
    \caption{Level 1 and 2 AP values obtained at each step of the experimental study in comparison with original the Faster R-CNN baseline.}
    \label{fig:summary_res}
\end{figure}

Figure \ref{fig:summary_res} summarizes the results obtained at each step of the experimental study in an incremental manner. As can be observed, single models already obtain a significant enhancement over the baseline, especially on pedestrian detection which is one of the underrepresented classes. The plot also illustrates the importance of the anchor generation optimization given the particularities of the object detection task in the autonomous driving context. Finally, the results obtained with the ensemble model demonstrate the effectiveness of combining different learning strategies in this complex scenario.

\subsection{Computational efficiency}

{In the autonomous driving scenario, the required computational resources and the inference speed are essential aspects to consider given that real-time predictions have to be provided in order to make informed driving decisions. Therefore, this section analyzes the computational efficiency of the studied models. Table} \ref{tab:times} {presents the training and inference computation time of the Faster R-CNN model, together with the other one-stage detectors that have been tested. The reported times are averaged over 1000 images with the original resolution and considering a batch size of one. Furthermore, other metrics are reported such as the number of parameters, the number of floating-point operations, and memory usage. For all tests, a computer with an Intel Core i7-770K CPU and a NVIDIA Titan V 12GB GPU has been used.}

First of all, it must be noted that the improvements proposed in this study do not increase the computation for training and inference compared to the original Faster R-CNN model with ResNet-101. The only extra cost of our proposal comes from the spatial features concatenation in the second stage, but it is minimal. The anchor optimization and the change in the training procedure do not introduce any computation overhead when running the detection model. {The anchor optimization process, which includes the clustering and genetic algorithm, requires around one hour of computation, but this is a step previous to the training of the model.}

As can be seen in Table \ref{tab:times}, in the experiments that use a single Faster R-CNN model, the training process is costly. It takes about 3 days to complete the $500k$ training steps that have been defined for the experiments. This illustrates the importance of using transfer learning, given the high cost of training this framework with very deep convolutional networks. With respect to the inference time, the Faster R-CNN model achieves a rate of 10.39 frames per second, which results in a practical object detection system in terms of speed \cite{Ren:2017}. Given that the Waymo data comes at 10 FPS, the proposed two-stage detector meets the real-time inference requirement. Furthermore, it can be seen that in the case of the ensemble, the overhead of the NMS algorithm is very small. The limitation of the ensemble is that as many devices as models will be needed to be efficient. If there are sufficient devices, predictions can be obtained in parallel for each model and combined very fast using the NMS algorithm.

{In comparison with the one-stage detectors, Faster R-CNN presents a faster inference speed than RetinaNet and CenterNet, but it is slower than YOLOv3. Although RetinaNet and CenterNet have lower memory usage, the use of feature pyramid networks in their backbone networks increases the computation time, making them less practical for this application. Moreover, the high resolution of the images is another factor that increases the inference time of these models. Although one-stage are faster than two-stage detectors in theory, using larger images reduces the differences between them} \cite{Carranza:2021}. {The number of parameters is similar in all models, with CenterNet having the lowest number. YOLOv3 is able to run at 15 FPS and has a very low memory usage thanks to the efficient DarkNet-53 backbone. It also presents a much lower number of FLOPS, which is a measure independent of the employed hardware. However, its detection precision is much worse than the two-stage detector as it was seen in Section} \ref{baseline}.  {In summary, it can be concluded that the proposed Faster R-CNN detector provides the best balance between accuracy and speed when compared to the rest of the one-stage models studied. },

\begin{table}[H]
\centering
\caption{{Computational efficiency of the studied detection models.}}
\label{tab:times}
\begin{tabular}{ccccccc}
\hline
\textbf{Model} & \textbf{Training (ms)} & \textbf{Inference (ms)} & \textbf{FPS} & \textbf{Parameters} $\mathbf{(10^6)}$ & \textbf{GFlops} & \textbf{Memory (GB)} \\ \hline
RetinaNet & 601.02 & 131.60 & 7.59 & 50.76 & 1310.64 & 3.53 \\
CenterNet & 586.68 & 103.52 & 9.65 & 44.37 & 919.11 & 3.12 \\
YOLOv3 & 320.34 & 65.12 & 15.35 & 55.63 & 418.26 & 2.35 \\ \hline
Faster R-CNN & 453.22 & 96.32 & 10.39 & 47.38 & 905.09 & 3.88 \\
+ NMS Ensemble & - & +0.15 & 10.38 & - & - & - \\ \hline
\end{tabular}
\end{table}

\section{Conclusions}

{In this paper, we proposed an enhanced 2D object detector based on Faster R-CNN that improves the detection accuracy in the context of autonomous driving.} In this scenario, the perspective projection plays an important role in the size of objects captured by on-board cameras. Furthermore, there is a high imbalance between vehicles and other traffic participants such as pedestrians and cyclists. Therefore, our study aimed to improve two main aspects: the default anchor generation procedure, and the performance drop in minority classes. We conducted a comprehensive experimental study over the Waymo Open Dataset, in which the proposed modifications over the original Faster R-CNN model were incrementally evaluated. 

After analyzing the properties of the objects in the dataset, we discovered a strong correlation between their size and their position in the image. Therefore, we proposed to improve the generation of the base anchors, which are a critical element for the success of the detector. {Our perspective-aware proposal was based on dividing the images into regions via clustering and searching for the optimal values for each of them using an evolutionary algorithm.} The improvement in performance obtained with the optimized parameters demonstrated that the default uniform anchors of Faster R-CNN were not suitable for this object detection task. {Moreover, given the importance of the spatial properties of proposals, we proposed to add a module to the second-stage header network that includes spatial information from the first-stage candidate regions to further increase localization accuracy.}

In order to address the {foreground-foreground} class imbalance, we explored different learning strategies that re-weight the original loss function. We found out that a modified version of focal loss can considerably enhance the detection of minority instances in two-stage detectors. Furthermore, the results showed that assigning weights according to the class distribution was more effective when dealing with very extreme class imbalance. Finally, we designed an ensemble model based on non-maximum suppression that combined the different training procedures. The ensemble model obtained a very significant increase in accuracy compared to the original Faster R-CNN baseline. 

In future works, we aim to study the application of our proposal under more efficient backbone networks to improve the inference rates of the detection system. {In the autonomous driving scenario, achieving real-time speed is crucial and should be given more importance in future studies. Moreover, another important line of work that should be addressed is the fusion of camera data with other sources, such as the information provided by LiDAR sensors. Novel approaches on how to combine the data of multi-modal sensors without damaging inference speed could lead to more robust and effective detectors.} {Further research will also study dynamic approaches that can guide the anchor generation to fit the data distribution during the training procedure, which could make the whole anchor optimization process automatic.} Furthermore, other ensemble techniques, such as graph-clique, could be tested and compared to the non-maximum-suppression approach.


\section*{Funding}{This research has been funded by FEDER/Ministerio de Ciencia, Innovación y Universidades – Agencia Estatal de Investigación/Proyecto TIN2017-88209-C2 and by the Andalusian Regional Government under the projects: BIDASGRI:  Big~Data technologies for Smart Grids (US-1263341), Adaptive hybrid models to predict solar and wind renewable energy production (P18-RT-2778).}

\section*{Acknowledgments}{We are grateful to NVIDIA
for their GPU Grant Program that has provided us the high-quality GPU devices for carrying out the study.}



\bibliographystyle{elsarticle-num} 
\bibliography{bibliography}

\end{document}